\documentclass[twoside,11pt]{article}
\usepackage{hyperref}
\hypersetup{ hidelinks }
\usepackage[abbrvbib, nohyperref]{jmlr2e}

\usepackage{multirow}
\usepackage{setspace}
\usepackage{amsmath}
\usepackage{amssymb}
\usepackage{bm}
\usepackage{placeins}
\usepackage{algorithm}
\usepackage{algorithmic}
\usepackage{lipsum}  
\usepackage[dvipsnames]{xcolor}
\usepackage{marvosym}
\usepackage{latexsym}
\usepackage{pifont}
\usepackage{diagbox}
\usepackage{url}            
\usepackage{amsfonts}       
\usepackage{nicefrac}       
\usepackage{wrapfig}
\usepackage{wasysym}
\usepackage{transparent}

\usepackage{microtype}
\usepackage{graphicx}
\usepackage{subfigure}
\usepackage{booktabs} 
\usepackage{enumitem}

\newcommand{\algcaption}[1]{%
  \refstepcounter{algorithm}%
  \hrule height.8pt depth0pt \kern2pt
  \textbf{Algorithm~\thealgorithm} #1
  \kern2pt\hrule\kern2pt
}

\usepackage{lastpage}
\jmlrheading{23}{2022}{1-\pageref{LastPage}}{10/21; Revised
3/22}{6/22}{21-1251}{Kaichao You, Yong Liu, Ziyang Zhang, Jianmin Wang, Michael I.\ Jordan, Mingsheng Long}

\ShortHeadings{Ranking and Tuning Pre-trained Models: A New Paradigm for Exploiting Model Hubs}{You, Liu, Zhang, Wang, Jordan, Long}
\firstpageno{1}

\begin{document}

\title{Ranking and Tuning Pre-trained Models: \\ A New Paradigm for Exploiting Model Hubs}

\author{\name Kaichao You$^{1 \; *}$ \email youkaichao@gmail.com \\
  \name Yong Liu$^1$ \thanks{The first two authors contributed equally to the paper.} \email liuyong21@mails.tsinghua.edu.cn \\
  \name Ziyang Zhang$^2$ \email zhangziyang11@huawei.com\\
  \name Jianmin Wang$^1$ \email jimwang@tsinghua.edu.cn\\
  \name Michael I.\ Jordan$^3$ \email jordan@cs.berkeley.edu \\
  \name Mingsheng Long$^1$ \thanks{Mingsheng Long is the corresponding author.} \email mingsheng@tsinghua.edu.cn \\
  \addr $^1$ School of Software, BNRist, Tsinghua University, Beijing 100084, China. \\
  \addr $^2$ Advanced Computing and Storage Lab, Huawei Technologies Co. Ltd \\
  \addr $^3$ Division of Computer Science and Department of Statistics, UC Berkeley, CA 94720-1776, USA
}

\editor{Stefan Harmeling}

\maketitle

\begin{abstract}
  Model hubs with many pre-trained models (PTMs) have become a cornerstone of deep learning. Although built at a high cost, they remain \emph{under-exploited}---practitioners usually pick one PTM from the provided model hub by popularity and then fine-tune the PTM to solve the target task. This na\"ive but common practice poses two obstacles to full exploitation of pre-trained model hubs: first, the PTM selection by popularity has no optimality guarantee, and second, only one PTM is used while the remaining PTMs are ignored. An alternative might be to consider all possible combinations of PTMs and extensively fine-tune each combination, but this would not only be prohibitive computationally but may also lead to statistical over-fitting. In this paper, we propose a new paradigm for exploiting model hubs that is intermediate between these extremes.  The paradigm is characterized by two aspects: (1) We use an evidence maximization procedure to estimate the maximum value of label evidence given features extracted by pre-trained models.  This procedure can rank all the PTMs in a model hub for various types of PTMs and tasks \emph{before fine-tuning}. (2) The best ranked PTM can either be fine-tuned and deployed if we have no preference for the model's architecture or the target PTM can be tuned by the top $K$ ranked PTMs via a Bayesian procedure that we propose. This procedure, which we refer to as \emph{B-Tuning}, not only improves upon specialized methods designed for tuning homogeneous PTMs, but also applies to the challenging problem of tuning heterogeneous PTMs where it yields a new level of benchmark performance.
\end{abstract}


\section{Introduction}

Deep neural networks~\citep{he_delving_2015, he_deep_2016,devlin_bert:_2019} trained on large-scale datasets~\citep{deng_imagenet:_2009, russakovsky_imagenet_2015,merity_pointer_2017} and specialized computational devices~\citep{jouppi_-datacenter_2017} have achieved striking, human-level performance on many pattern recognition tasks in both computer vision and natural language processing. Moreover, research has shown that deep neural networks trained on large-scale pre-training tasks~\citep{yang_xlnet:_2019,clark_electra:_2020,brown_language_2020} can produce generic representations~\citep{donahue_decaf:_2014} that benefit downstream tasks such as object detection~\citep{girshick_rich_2014} and language understanding~\citep{wang_superglue:_2019}. These trained neural networks are known as pre-trained models (PTMs). Readers can refer to dedicated surveys~\citep{han_pre-trained_2021,qiu_pre-trained_2020,bommasani_opportunities_2021} for a holistic overview of pre-trained models. The power~\citep{brown_language_2020} of PTMs, together with the transfer learning paradigm of ``pre-training $\rightarrow$ fine-tuning'' to exploit PTMs, has had significant impact in both vision~\citep{kornblith_better_2019} and language~\citep{devlin_bert:_2019}, and the influence of PTMs is growing in nearby communities such as geometric learning~\citep{hu_strategies_2020}.

The cost of training PTMs varies from hundreds of GPU hours~\citep{he_deep_2016} to hundreds of GPU days~\citep{devlin_bert:_2019}, which can be prohibitively expensive for individual researchers and academic labs. Most pre-trained models are provided by central repositories, including PyTorch Hub,\footnote{\url{https://pytorch.org/hub/}} TensorFlow Hub,\footnote{\url{https://www.tensorflow.org/hub}} and HuggingFace Transformer Models.\footnote{\url{https://huggingface.co/models}} Such collections of pre-trained models are called a ``pre-trained model hubs'' (PTM hubs), and they have become very popular; for example, the HuggingFace Transformer library~\citep{wolf_transformers:_2020}, the most popular BERT model~\citep{devlin_bert:_2019} is currently being downloaded over $80$ million times every month.

\begin{figure}[tbp]
  \includegraphics[width=.7\columnwidth]{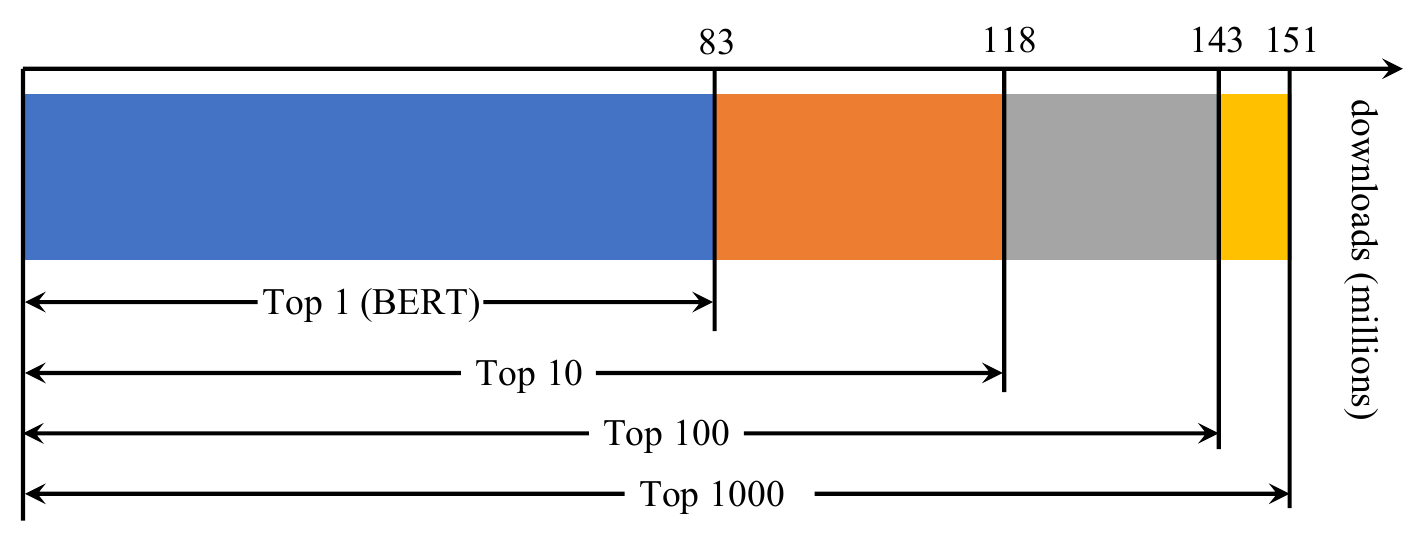}
  \centering
  \vspace{-10pt}
  \caption{Monthly download statistics (in millions) for top popular models in the HuggingFace Transformer library. The most popular PTM (BERT) takes up more than a half downloads, while other PTMs are much less exploited.}
  \label{fig:download}
\end{figure}

Although centralized repositories spend enormous resources to provide large-scale PTM hubs to the public, it turns out that practitioners often pick the most popular PTM, meaning that the whole PTM hub is insufficiently exploited. Figure~\ref{fig:download} analyzes the monthly downloads of PTMs in the HuggingFace Transformer hub. Beyond  several popular models the remaining PTMs in the hub are seldom downloaded. The statistics in PyTorch Hub and TensorFlow Hub are essentially the same---several popular PTMs dominate the rest.

Na\"ively picking the most popular PTM is far from optimal in two respects: (1) The PTM selection is \emph{task-specific} and one PTM cannot be optimal for all the tasks; different tasks generally favor different PTMs, depending on the compatibility between the pre-trained model and the target task~\citep{you_logme:_2021}. (2) Only one PTM is exploited, and opportunities to obtain benefits from ensembling or aggregation are put aside. Correspondingly, there are two reasons why practitioners resort to the suboptimal na\"ive practice: (1) maximally exploiting a PTM hub requires trying all combinations of PTMs and extensively fine-tuning each PTM combination, which requires unaffordable computation; (2) even if the significant computational cost can be paid, it is unclear how to exploit multiple PTMs in transfer learning. As discussed in Section~\ref{sec:multi_transfer}, \citet{shu_zoo-tuning:_2021} studied the problem in a limited case, but a general solution is lacking.

\begin{figure}[tbp]
  \centering
  \includegraphics[width=.9\columnwidth]{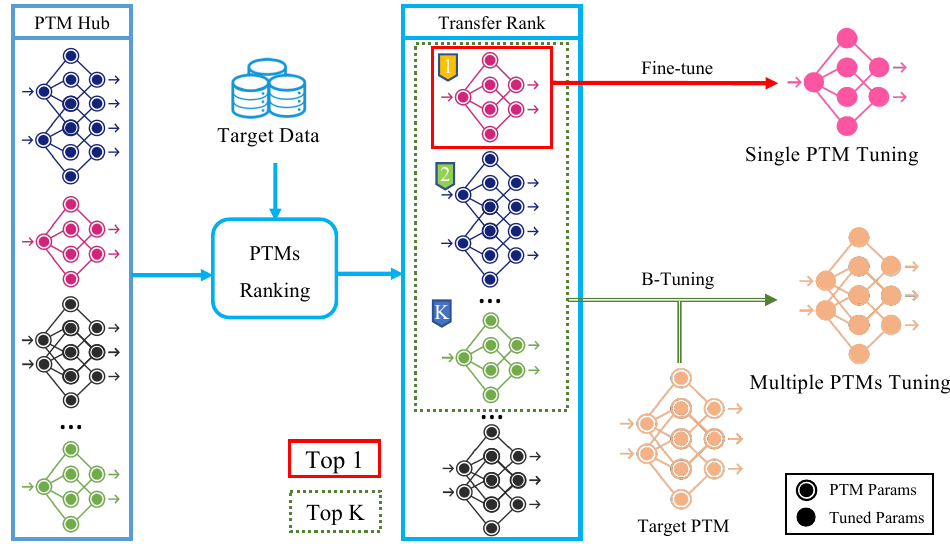}
  \caption{The proposed paradigm of ranking and tuning pre-trained models. PTMs are ranked by their transferability w.r.t.~the target data, then either the best PTM is fine-tuned, or the target PTM is tuned by top-K PTMs via the proposed B-Tuning.}
  \label{fig:overview}
\end{figure}

To fully exploit PTM hubs, we propose a new paradigm: ranking and tuning pre-trained models. Figure~\ref{fig:overview} provides an overview of the paradigm. It consists of two parts: (1) PTMs are \emph{ranked} by a transferability metric; (2) top-ranked PTMs are \emph{tuned} to meet downstream applications' requirements. Our preliminary work~\citep{you_logme:_2021} proposed a method that we referred to as ``logarithm of maximum evidence'' (LogME) to estimate the compatibility between PTMs and downstream datasets. We demonstrated its effectiveness on a variety of PTMs. With an effective transferability rank, the best ranked PTM can be fine-tuned if there are no constraints on network architecture like inference time or hardware-friendly operators. If these constraints are present, the qualified PTM with desired architecture might not be the best-ranked one, but it can be tuned by top-K ranked PTMs via a novel B-Tuning algorithm as discussed in Section~\ref{sec:bayesian_tuning}.

Compared with picking the most popular PTM, our proposed new paradigm features two significant advantages: (1) it provides a \emph{task-adaptive ranking} of all PTMs in a PTM hub to enable optimal selection of PTMs; (2) it opens the new possibility to \emph{exploit multiple PTMs for tuning}, breaking the stereotype that fine-tuning must be tied up with a single PTM. The new paradigm can be useful in a broad variety of scenarios, as pre-trained models are increasingly important in deep learning.

Besides a new paradigm of exploiting PTM hubs, this paper brings novel theoretical analyses and a new algorithm for multiple PTM tuning. (1) On the theoretical side, we derive a sufficient condition for the evidence maximization algorithm~\citep{mackay_bayesian_1992} to converge and analyze the influence of dimensionality on LogME. The evidence maximization algorithm~\citep{mackay_bayesian_1992} has been used primarily as a heuristic; a rigorous convergence condition has been lacking. (2) On the algorithm design side, we devise a method that we refer to as ``B-Tuning'' for tuning multiple PTMs using Bayesian learning.  We show that this method surpasses the dedicated method~\citep{shu_zoo-tuning:_2021} for homogeneous PTMs (PTMs with the same architecture) and also works for the challenging scenario with heterogeneous PTMs (PTMs with different architectures).

The contributions of this paper are summarized as follows:

\begin{enumerate}[noitemsep,topsep=0pt]
  \item We propose a new paradigm for exploiting PTM hubs, namely ranking and tuning pre-trained models. It has significant advantages compared with the common practice of na\"ively fine-tuning a popular pre-trained model.
  \item Concerning ranking PTMs, we propose LogME for transferability assessment and develop a fast algorithm to accelerate the computation. LogME is easy to interpret and is extremely efficient: it brings roughly $3700\times$ speedup in wall-clock time and requires just $1\%$ memory footprint compared with brute-force fine-tuning. Theoretical analyses confirm the rationality of LogME, and lay a theoretical foundation for a heuristic algorithm in evidence maximization.
  \item For tuning PTMs, two possible scenarios are studied. In the academic scenario without specific requirements for the PTM architecture, the best ranked pre-trained model according to the transferability rank can be selected for subsequent fine-tuning; in the industrial scenario where a specific PTM architecture is required to meet the budget of computation and energy, we propose B-Tuning to tune the given pre-trained model with top-K ranked PTMs, even though these PTMs are heterogeneous.
\end{enumerate}

Compared with our conference paper~\citep{you_logme:_2021} that only proposed LogME for transferability estimation, this paper extends LogME to a paradigm of ranking and tuning pre-trained models. Additional theoretical analyses are available in the ranking part, and a new algorithm is presented in the tuning part. Moreover, LogME is tested against additional tasks like named entity recognition~\citep{sang_introduction_2003} in Section~\ref{sec:ner_experiments} and prompt learning~\citep{liu_pre-train_2021} in Section~\ref{sec:compare_head}. 

We will need to make significant use of notation to describe our ideas; thus, for the convenience of readers, all of the notation is collected in Table~\ref{tab:notations}.

The basic problem setup contains a PTM hub with $M$ pre-trained models, $\{\phi_k\}_{k=1}^{M}$, and the transfer learning task is given by a labeled dataset, $\mathcal{D} = \{(x_i, Y_i)\}_{i=1}^{n}$, with $n$ labeled data points. This paper focuses on classification and regression tasks, so the label $Y_i \in \mathbb{R}^C$ is $C$ dimensional. 

The rest of the paper is organized as follows: Section~\ref{sec:related} summarizes related work, Section~\ref{sec:ranking_ptms} focuses on ranking and describes the LogME transferability metric, Section~\ref{sec:theoretical} presents theoretical analyses for LogME, Section~\ref{sec:tuning_ptms} focuses on tuning and introduces the B-Tuning method for multiple PTM tuning, Section~\ref{sec:experiments} presents all the experiments, and finally Section~\ref{sec:conclusion} concludes the paper.

\section{Related Work}
\label{sec:related}

\subsection{Transfer learning}
\label{sec:transfer}

Transfer learning~\citep{thrun_learning_1998} consists of transductive transfer, inductive transfer, task transfer, and so on. A well-known transductive paradigm is domain adaptation~\citep{quionero-candela_dataset_2009}, which aims for reducing domain shifts by transferring samples, hidden features~\citep{long_learning_2015, ganin_unsupervised_2015}, and categorical information~\citep{cao_big_2022}. Inductive transfer, in particular fine-tuning in deep learning~\citep{erhan_why_2010, yosinski_how_2014}, exploits prior knowledge (pre-trained models) to improve the performance of target tasks. Task transfer learning~\citep{zamir_taskonomy:_2018} focuses on how to transfer tasks rather than pre-trained models. It aims to discover shared relevance among tasks~\citep{ben-david_exploiting_2003} and to exploit the relationship for improvement on the target task. In the context of deep learning, transfer learning usually means inductive transfer with a pre-trained model, which is the focus of this paper.

Many previous works~\citep{yosinski_how_2014,kornblith_better_2019,neyshabur_what_2020} have shown the benefit of initializing a deep neural network with a pre-trained model. Apart from the vanilla method (\emph{i.e.}, the pre-trained model is just used for initialization), researchers have recently proposed sophisticated fine-tuning techniques like regularization~\citep{li_explicit_2018,chen_catastrophic_2019}, additional supervision~\citep{you_co-tuning_2020}, and carefully designed architectures~\citep{kou_stochastic_2020}. They can further improve transfer learning performance, but empirically \emph{these fine-tuning methods do not change the ranking of pre-trained models on downstream tasks}. That is, if pre-trained model A is better than pre-trained model B after vanilla fine-tuning, empirically A is better than B when those advanced techniques are integrated. For example, on three datasets and four sampling rates in Table~2 of \citet{you_co-tuning_2020}, better fine-tuning performance primarily indicates better Co-Tuning (their proposed method) performance, implying that the transferability of a pre-trained model might be \emph{task-specific} rather than \emph{method-specific}. Therefore, our experiments stick to the vanilla fine-tuning during PTM ranking.

\subsection{PTMs and PTM hubs}

Pre-trained models (PTMs) are generalizable deep networks trained on large-scale data. They can be transferred to a series of downstream tasks. They have become a cornerstone in deep learning and sometimes are known as foundation models~\citep{bommasani_opportunities_2021}. Typical categories of PTMs are summarized in the following.

\textbf{Supervised PTMs.} In the ImageNet classification challenge, \citet{he_delving_2015} developed the first deep neural network that surpassed human performance. By supervised pre-training on the ImageNet dataset, deep models marched towards higher accuracy, fewer parameters, and lower computation. InceptionNet~\citep{szegedy_going_2015} made use of parallel convolutional filters to extract different levels of features. ResNet~\citep{he_deep_2016} introduced skip-connections to ease the vanishing gradient problem so that much deeper networks could be trained. Inspired by ResNet, DenseNet~\citep{huang_densely_2017} was equipped with dense skip-connections to reuse features in a parameter-efficient manner. MobileNet~\citep{sandler_mobilenetv2:_2018} was a low-parameter, mobile-friendly network structure which was further optimized with the help of network architecture search to become MNASNet~\citep{tan_mnasnet:_2019}.

\textbf{Unsupervised PTMs.} Although supervised pre-training is the most common practice, the labeling cost of large-scale data can be prohibitively expensive. As a large amount of unlabeled data on the Internet are available but under-exploited, recently many researchers have sought to apply self-supervised learning~\citep{jing_self-supervised_2020} on unlabeled data~\citep{mahajan_exploring_2018} with contrastive loss~\citep{gutmann_noise-contrastive_2010}. Accordingly, a family of unsupervised deep models has emerged in recent years. \citet{he_momentum_2020} proposed Momentum Contrast with a creative queue structure to fully exploit the manifold structure of unlabeled data. \citet{chen_simple_2020} significantly improved performance by exploring data augmentation, multi-layer projection head, and empirical designs. Designing better strategies for contrastive pre-training is still under active research~\citep{tian_what_2020}.

\textbf{Language PTMs.} In recent years, natural language processing has been revolutionized by language PTMs. Unsupervised pre-trained models have been well established by training masked language models~\citep{devlin_bert:_2019} or autoregressive language models~\citep{yang_xlnet:_2019} on large unlabeled corpora~\citep{merity_pointer_2017}. \citet{liu_roberta:_2019} explored many practical aspects of the training of language models. \citet{sanh_distilbert_2019} proposed distillation to make PTMs smaller and faster. These pre-trained language models are very common in winning submissions on important benchmarks like GLUE~\citep{wang_glue:_2018} and SQuAD~\citep{rajpurkar_squad:_2016}, and have established their profound influence in the industry.

PTMs are grouped together to be hosted in PTM hubs like \href{https://pytorch.org/vision/stable/models.html}{TorchVision} and \href{https://huggingface.co/models}{HuggingFace Models}. Industry labs have invested significant resources in training these PTMs, but unfortunately, PTM hubs are under-exploited, as quantitatively measured in Figure~\ref{fig:download} and described in the introduction section. The goal of this paper is to develop a new paradigm of exploiting PTM hubs, so that pre-trained models can be more widely exploited.

\subsection{Assessing the transferability of pre-trained models}

Assessing the transferability of PTMs has great significance in guiding the practice of deep learning. It can be used to rank available PTMs and act as a criterion for pre-trained model selection. \citet{yosinski_how_2014} studied the performance of transferring different layers of a pre-trained model, and \citet{kornblith_better_2019} studied a wide variety of ImageNet PTMs with modern network architectures. These papers aim for a deep understanding~\citep{neyshabur_what_2020} of transfer learning by expensive and exhaustive fine-tuning with major computation cost (see Section~\ref{sec:timecost}), which is hard for practitioners to afford. In most scenarios, practitioners care most about PTMs' relative ranking on target tasks to guide PTM selection, requiring a practical assessment method that is \emph{efficient}, \emph{accurate}, and \emph{general}: a transferability assessment method should be efficient enough compared with brute-force fine-tuning~\citep{zamir_taskonomy:_2018}, should be accurate enough to identify potentially best models, and should be general enough to tackle a wide variety of common scenarios.

LEEP~\citep{nguyen_leep:_2020} and NCE~\citep{tran_transferability_2019} were the first two methods to assess the transferability of pre-trained models. \citet{nguyen_leep:_2020} constructed an empirical predictor from the joint distribution $p(y_t, y_s)$ over pre-trained labels $y_s$ and target labels $y_t$, and calculated the log expectation of the empirical predictor (LEEP) as the transferability measure. The empirical predictor predicts the probability of the target class $y_t$ as $\sum_{y_s \in \mathcal{Y}_s} p(y_t | y_s) p(y_s)$, where $p(y_s)$ comes from the PTM's prediction over pre-trained categories. Negative Conditional Entropy (NCE) proposed by \citet{tran_transferability_2019} depended on an information-theoretic quantity~\citep{cover_elements_1999} to reveal the transferability and hardness between different tasks. It estimated the joint distribution $p(y_t, y_s)$ with one-hot labels and predictions, and defined NCE as $-H(y_t | y_s)$, \emph{i.e.}, the negative conditional entropy of target labels $y_t$ given PTM's predictions $y_s$.

\begin{table}[htbp]
  \centering
  \caption{Applicability of existing methods and LogME proposed in this paper.}
  \vskip 0.05in
    \begin{tabular}{cccccc}
      \toprule
      \multirow{2}[0]{*}{Modality} & \multirow{2}[0]{*}{Pre-train} & \multirow{2}[0]{*}{Target} & \multicolumn{3}{c}{Method} \\
      \cline{4-6}
      & & & LEEP & NCE & LogME \\
    \midrule
    \multirow{4}[0]{*}{vision} & classification & classification &   \ding{51}    &   \ding{51}    & \ding{51} \\
          & classification & regression &  {\transparent{0.5} \ding{55}}     &  {\transparent{0.5} \ding{55}}     & \ding{51} \\
          & contrastive & classification &       {\transparent{0.5} \ding{55}}     &  {\transparent{0.5} \ding{55}}     & \ding{51}  \\
          & contrastive & regression &     {\transparent{0.5} \ding{55}}     &  {\transparent{0.5} \ding{55}}     & \ding{51} \\
          \midrule
    language & language modeling & classification &       {\transparent{0.5} \ding{55}}     &  {\transparent{0.5} \ding{55}}     & \ding{51}  \\
    \bottomrule
    \end{tabular}
  \label{tab:applicability}%
\end{table}%

All of these methods, however, had their limitations. As shown in Table~\ref{tab:applicability}, they can only handle classification tasks with supervised pre-trained models. Increasingly popular contrastive pre-trained models and language models are out of their scope. The LogME algorithm proposed in this paper extends the applicability of transferability assessment to these cases. LogME is fast to compute, less prone to over-fitting, and broadly applicable to various pre-trained models/downstream tasks/data modalities. Its performance is validated by extensive experiments. Prior to this paper, for most (four out of five) transfer learning settings, task adaptive transferability assessment did not have a satisfactory solution. In addition, LogME's statistical rigor makes it extensible to multiple PTM tuning (see Section~\ref{sec:bayesian_tuning}), which further fleshes out the new paradigm of ranking and tuning pre-trained models.

\subsection{Multiple PTM tuning}
\label{sec:multi_transfer}

A straightforward approach in transfer learning is to fine-tune models initialized from pre-trained parameters, which we call ``\emph{single PTM tuning}'' because it can only exploit a specific pre-trained model during fine-tuning.

It is widely acknowledged that the success of transfer learning comes from the knowledge in the pre-trained model. Considering that there are so many PTMs in a PTM hub, it is appealing to transfer multiple PTMs simultaneously, a problem we call ``\emph{multiple PTM tuning}.'' We might expect multiple PTM tuning to outperform single PTM tuning. 

Unfortunately, multiple PTM tuning is under-explored due to  technical challenges. If multiple PTMs are homogeneous, \emph{i.e.}, they share the same network architecture, the problem becomes easier. Researchers in this area focused on how to align and merge multiple homogeneous PTMs. \citet{singh_model_2020} defined transportation cost between neural representations and minimized the induced Wasserstein distance to align neurons from each PTM. \citet{shu_zoo-tuning:_2021} developed a channel-wise alignment method dedicated to convolutional neural networks with a learnable gating function to merge multiple PTMs. Prior to this paper, \citet{shu_zoo-tuning:_2021} held the state-of-the-art result in homogeneous PTMs tuning.

Heterogeneous PTMs tuning is much more difficult than homogeneous PTMs tuning and it is still unresolved how to address this general form of tuning. In practice, pre-trained models in PTM hubs generally have different architectures  and it has become increasingly urgent to address the heterogeneity problem.

This paper proposes a methodology for exploiting general PTM hubs. In the proposed paradigm, PTMs are first ranked by LogME, then the top-K ranked PTMs from the PTM hub are selected for multiple PTM tuning. A Bayesian tuning method (B-Tuning, see Section~\ref{sec:bayesian_tuning}) is further proposed to solve the multiple PTM tuning problem. Overall the method proposes a solution to the heterogeneous PTMs tuning problem, and it surpasses the state-of-the-art method~\citep{shu_zoo-tuning:_2021} dedicated to homogeneous PTM tuning.

\section{Ranking Pre-Trained Models}
\label{sec:ranking_ptms}

The ranking of pre-trained models requires a transferability metric. But before introducing the transferability metric in Section~\ref{sec:logme}, we discuss how to quantify its fidelity to the reference transferability performance, which is elaborated in the following Section~\ref{sec:how_to_measure_ranking}.

\subsection{How to measure the performance of a transferability metric?}
\label{sec:how_to_measure_ranking}

A transfer learning task (in the form of a dataset $\mathcal{D} = \{(x_i, Y_i)\}_{i=1}^{n}$) should have an evaluation metric (accuracy, MAP, MSE, \emph{etc.}) to measure the reference transfer performance $T_k$ of fine-tuning $\phi_k$ with sufficient hyper-parameter tuning. A practical assessment method should produce a score $S_k$ for each pre-trained model $\phi_k$ (ideally without fine-tuning $\phi_k$ on $\mathcal{D}$), and the scores $\{S_k\}_{k=1}^{M}$ should well correlate with $\{T_k\}_{k=1}^{M}$ so that top-performing pre-trained models can be selected by simply evaluating the scores $\{S_k\}_{k=1}^{M}$.

A perfect pre-trained model assessing method would produce $\{S_k\}_{k=1}^{M}$ with precisely the same order as $\{T_k\}_{k=1}^{M}$. To measure the deviation from the perfect method, we can use simple metrics like top-1 accuracy or top-K accuracy (whether the fraction among top-K in $\{S_k\}_{k=1}^{M}$ are also top-K in $\{T_k\}_{k=1}^{M}$). Nevertheless, top-1 accuracy is too conservative and top-K accuracy is not comparable across different values of $M$. Rank correlation~\citep{fagin_comparing_2003} is a good alternative to directly measure the correlation between $\{S_k\}_{k=1}^{M}$ and $\{T_k\}_{k=1}^{M}$. Prior work~\citep{nguyen_leep:_2020} adopted Pearson's linear correlation coefficient, but neither Pearson's linear correlation nor its variant (Spearman's rank correlation) has a simple interpretation (see the interpretation of $\tau$ below). Therefore, they are not used in this paper.

The rank correlation method we choose is Kendall's $\tau$ coefficient~\citep{kendall_new_1938}, which counts concordant pairs to capture the possibility of \emph{$T_i$ being better than $T_j$ if $S_i$ is better than $S_j$} in choosing a good pre-trained model.

Without loss of generality, we assume larger values of transfer performance $T$ and score $S$ are preferred (\emph{e.g.}, accuracy). If this is not the case (\emph{e.g.}, transfer performance is measured by mean square error and small values are favored), the negation $-T$ can be considered. For a pair of measures $(T_i, S_i)$ and $(T_j, S_j)$, the pair is concordant if $T_i < T_j \land S_i < S_j$ or $T_i > T_j \land S_i > S_j$ (concisely speaking, $\text{sgn}(T_i - T_j) \text{sgn}(S_i - S_j) = 1$).  Kendall's $\tau$ coefficient is defined by the following equation, which enumerates all $\binom{M}{2}$ pairs and counts the number of concordant pairs minus the number of discordant pairs.
\begin{equation*}
    \tau = \frac{\sum_{1 \le i<j \le M} \text{sgn}(T_i - T_j) \text{sgn}(S_i - S_j) }{\binom{M}{2}}.
\end{equation*}

\textbf{How to interpret $\tau$}~\citep{fagin_comparing_2003}: The range of $\tau$ is $[-1, 1]$. $\tau=1$ means $T$ and $S$ are perfectly correlated ($S_i > S_j \iff T_i > T_j$), and $\tau=-1$ means $T$ and $S$ are inversely correlated ($S_i > S_j \iff T_i < T_j$). \emph{If $T$ and $S$ have a correlation value of $\tau$, the probability of $T_i > T_j$ is $\frac{\tau + 1}{2}$ when $S_i > S_j$}.

\textbf{Pay attention to top-performing models.} Since a major application of transferability metric is to select top-performing pre-trained models, discordant/concordant pairs should be weighted more if $T_i, T_j, S_i, S_j$ are larger. This can be taken care of by $\tau_w$~\citep{vigna_weighted_2015}, a weighted variant of Kendall's $\tau$. The details of calculating $\tau_w$ can be found in the SciPy \href{https://docs.scipy.org/doc/scipy/reference/generated/scipy.stats.weightedtau.html}{implementation}. With the weighting scheme, correlation value $\tau_w$ corresponds to a proportion interval of concordant pairs rather than a unique proportion value $\frac{\tau_w + 1}{2}$. Nevertheless, the interval lies near the value $\frac{\tau_w + 1}{2}$. Therefore, we can roughly use the probability $\frac{\tau_w + 1}{2}$ of concordant pairs to interpret the correlation value $\tau_w$.

In short, we measure the correlation between $\{S_k\}_{k=1}^{M}$ and $\{T_k\}_{k=1}^{M}$ by $\tau_w$~\citep{vigna_weighted_2015}. \emph{Larger $\tau_w$ indicates a better correlation and better assessment.}


\subsection{The LogME approach}
\label{sec:logme}

This section describes LogME in detail. Since a transferability metric measures the transferability of pre-trained models, it should produce a score $S_k$ for each PTM $\phi_k$ independent of the remaining PTMs. We thus drop the subscript $k$ in this section.

An important goal of designing transferability metrics is to quickly assess many PTMs. With that in mind, we set the minimization of assessment time as a priority. First, to avoid expensive optimization of the whole PTM, PTM $\phi$ is regarded as a fixed feature extractor. Note that \citet{nguyen_leep:_2020} were limited to supervised pre-trained models because they used a pre-trained classification head $h$. In contrast, \emph{we only use the pre-trained representation model $\phi$ so that the proposed method can be applied to any pre-trained model} (whether supervised pre-trained or unsupervised pre-trained).

With $\phi$ fixed, features $\{f_i = \phi(x_i)\}_{i=1}^{n}$ and labels $\{Y_i\}_{i=1}^{n}$ of the target task are the ingredients we can use to assess pre-trained models. The rest of this section discusses how to estimate the compatibility of features and labels as a transferability metric.

\subsubsection{Evidence calculation}
\label{sec:evidence}

We first consider a simple case with $D$-dimensional features $f_i \in \mathbb{R}^D$ and scalar labels $y_i \in \mathbb{R}$. Note that the actual label $Y_i$ can be non-scalar, and the way in which we extend from scalar labels $y_i$ to vector labels $Y_i$ is explained in Section~\ref{sec:evidence_maximization}.

Let the feature matrix $F \in \mathbb{R}^{n \times D}$ denote all the features and $y \in \mathbb{R}^{n}$ denote all the labels. A direct measurement of the compatibility between features $F$ and labels $y$ is the probability density $p(y | F)$, which is intractable without a parameterized model. Since the rule-of-thumb transfer learning practice is to add a linear layer on top of the pre-trained model, we use a linear model upon features parameterized by $w$.

A straightforward approach to deal with the linear model is to find the best $w^*$ by logistic or linear regression under maximum likelihood estimation, and to assess pre-trained models by the likelihood $p(y | F, w^*)$. However, it is well known that \textit{maximum likelihood estimation is prone to over-fitting}~\citep{bishop_pattern_2006}. Regularization techniques like $\ell_2$-regularization may alleviate over-fitting at the cost of additional hyper-parameters, which requires manual intervention or grid search to tune those hyper-parameters. Even after extensive hyper-parameter tuning, its performance is not satisfying as observed in Section~\ref{sec:compare_head}, because finding an optimal hyper-parameter is very difficult. \emph{Ideally, a transferability metric should have no hyper-parameters so that it can be applied to downstream tasks without manual intervention.} Obviously, this approach does not satisfy the hyper-parameter-free property.

\begin{figure}[htbp]
  \centering
  \includegraphics[width=.6\columnwidth]{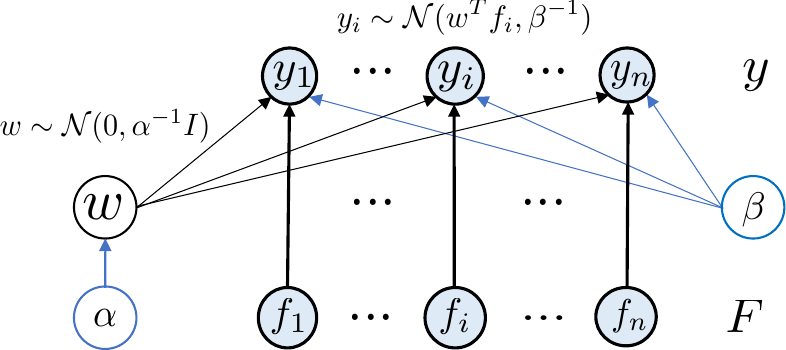}
  \caption{The directed graphical model for calculating evidence.}
  \label{fig:graph}
\end{figure}

The disadvantage of the above approach can be overcome by the evidence approach introduced below. Evidence (also known as marginalized likelihood) is defined as $p(y|F) = \int p(w) p(y | F, w) \mathrm{d} w$, which integrates over all possible values of $w$ rather than taking one $w^*$ value. This evidence-based approach is an elegant model selection approach and has a rigorous theoretical foundation~\citep{knuth_bayesian_2015}. $p(w)$ and $p(y | F, w)$ are modeled by a graphical model (Figure~\ref{fig:graph}) specified by two positive hyper-parameters $\alpha$ and $\beta$: the prior distribution of the weight is an isotropic multivariate Gaussian $w \sim \mathcal{N}(0, \alpha^{-1}I)$, and the distribution of each observation is a one-dimensional normal distribution $p(y_i | f_i, w, \beta) \sim \mathcal{N}(y_i | w^T f_i, \beta^{-1})$. Fortunately, hyper-parameters $\alpha$ and $\beta$ can be automatically set to their optimal values as described in Section~\ref{sec:evidence_maximization}.

According to the causal structure in Figure~\ref{fig:graph} and the basic principles in graphical models~\citep{koller_probabilistic_2009}, the evidence can be calculated analytically as follows:
\begin{equation}
  \label{eq:likelihood}
  \resizebox{0.94\textwidth}{!}{
    $ p(y|F, \alpha, \beta)
    = \int p(w | \alpha) \prod_{i=1}^{n} p(y_i | f_i, w, \beta) \mathrm{d} w
    = (\frac{\beta}{2 \pi})^{\frac{n}{2}} (\frac{\alpha}{2 \pi})^{\frac{D}{2}} \int e^{-\frac{\alpha}{2} w^Tw -\frac{\beta}{2} \vert \vert Fw - y \vert \vert^2}  \mathrm{d} w .$
  }
\end{equation}

Equation~\ref{eq:likelihood} can be simplified by the identity $\int e^{- \frac{1}{2} (w^TAw + b^Tw + c)} \, \mathrm{d}w = \sqrt{\frac{(2\pi)^D}{\vert A \vert}} e^{-\frac{1}{2}c + \frac{1}{8}b^TA^{-1}b}$. Taking the logarithm for simplicity, Equation~\ref{eq:evidence} shows the logarithm of the evidence $\mathcal{L}$ as a function of $\alpha, \beta$, where $A = \alpha I + \beta F^TF, m = \beta A^{-1}F^Ty$.
\begin{equation}
  \label{eq:evidence}
  \resizebox{0.94\textwidth}{!}{
  $\mathcal{L}(\alpha, \beta) =  \log p(y|F, \alpha, \beta) = \frac{n}{2} \log \beta  + \frac{D}{2} \log \alpha - \frac{n}{2} \log 2\pi - \frac{\beta}{2} \vert \vert F m - y \vert \vert_2^2 - \frac{\alpha}{2} m^Tm - \frac{1}{2} \log \vert A \vert $.
  }
\end{equation}

\subsubsection{Evidence maximization and LogME}
\label{sec:evidence_maximization}

An unresolved issue in Equation~\ref{eq:evidence} is how to choose $\alpha, \beta$. \citet{gull_developments_1989} suggested choosing $\alpha, \beta$ to maximize the evidence, \emph{i.e.}, use $(\alpha^*, \beta^*) = \arg \max_{\alpha, \beta} \mathcal{L}(\alpha, \beta)$. Because $m$ and $A$ are coupled, it is a difficult problem to directly maximize $\mathcal{L}(\alpha, \beta)$. To address this, \citet{mackay_bayesian_1992} proposed a heuristic algorithm to solve the maximization problem: (1) set initial value of $\alpha, \beta$; (2) evaluate $A, m, \gamma$ with given $\alpha, \beta$: $A = \alpha I + \beta F^TF, m = \beta A^{-1}F^Ty, \gamma = \sum_{i=1}^{D} \frac{\beta \sigma_i^2}{\alpha + \beta \sigma_i^2}$, where $\sigma_i$ are singular values of $F$; (3) maximize $\alpha, \beta$ by solving $\frac{\partial \mathcal{L}}{\partial \alpha} = 0, \frac{\partial \mathcal{L}}{\partial \beta} = 0$ with $m, \gamma$ fixed, which yields $\alpha \leftarrow \frac{\gamma}{m^Tm}, \beta \leftarrow \frac{n - \gamma}{\vert \vert F m - y \vert \vert_2^2}$. The algorithm is called MacKay's algorithm (Algorithm~\ref{alg:mackay_algorithm}). Section~\ref{sec:convergence_analysis} gives a theoretical convergence guarantee for the algorithm. Interestingly, the fixed point iteration used for the convergence analysis can also be used in practice to obtain a new and faster algorithm for evidence maximization (Algorithm~\ref{alg:optimized_fixed_point}). Please refer to Section~\ref{sec:convergence_analysis} for details.

After the convergence of evidence maximization, the logarithm maximum evidence $\mathcal{L}(\alpha^*, \beta^*)$ is used to evaluate the compatibility between features and labels. Because $\mathcal{L}(\alpha^*, \beta^*)$ scales linearly with $n$, we normalize it as $\frac{\mathcal{L}(\alpha^*, \beta^*)}{n}$ and term it \textbf{LogME} (logarithm of maximum evidence). Discussion on the influence of dimensionality $D$ is presented in Section~\ref{sec:influence_of_dimension}. LogME can be intuitively interpreted as the logarithm of maximum label evidence given pre-trained features.

\textbf{Extending LogME to complex cases.} The LogME approach starts from a single-target regression. If the target problem is a multivariate regression task, \emph{i.e.}, $Y \in \mathbb{R}^{n\times C}$, we can calculate LogME for each dimension $c \; (1 \le c \le C)$ and average them over the $C$ dimensions. If the target problem is a classification task with $C$ classes, Equation~\ref{eq:likelihood} cannot be calculated analytically~\citep{daunizeau_semi-analytical_2017} with a categorical prior distribution. State-of-the-art approximation methods like Laplace approximation~\citep{immer_scalable_2021} work well in toy data, but perform unsatisfyingly in realistic tasks, as mentioned later in Section~\ref{sec:toy}. Therefore, we turn to an alternative solution: convert the classification labels to one-hot labels and treat the problem as multivariate regression. This approach also works for multi-label classification. This way, LogME can be used in both (single-label and multi-label) classification and regression tasks.

The overall algorithmic specification of LogME is presented in Algorithm~\ref{alg:logme}.

\begin{figure}[htbp]
	\algcaption{LogME}
	\label{alg:logme}
	\begin{algorithmic}[1]
		\STATE {\bfseries Input:} Pre-trained model $\phi$ and target dataset $\mathcal{D} = \{(x_i, Y_i)\}_{i=1}^{n}$  \\ 
		\STATE {\bfseries Output:} Logarithm of Maximum Evidence (LogME)
				
		\STATE { Extract} features using pre-trained model $\phi$: $F \in \mathbb{R}^{n\times D}$, $f_i = \phi(x_i)$, $Y \in \mathbb{R}^{n\times C}$
    		
    \STATE{Compute} SVD of $F$: $F = U\Sigma V^T$. Then $F^TF = V\text{diag}\{\sigma^2\}V^T$ \\
    
    \FOR{\text{dimension} $c=1$ \text{to} $C$ } 
    \STATE{Let} $y = Y^{(c)} \in \mathbb{R}^n$, \\ 
    \STATE{Calculate} the LogME value $\mathcal{L}_c$ by evidence maximization (Algorithm~\ref{alg:mackay_algorithm} or Algorithm~\ref{alg:optimized_fixed_point}).
    \ENDFOR
		\STATE{Return} LogME $\frac{1}{C}\sum_{c=1}^{C} \mathcal{L}_c$ \\
	\end{algorithmic}
  \kern2pt\hrule\relax

  \bigskip
  \bigskip

	\algcaption{Evidence Maximization by MacKay's Algorithm}
	\label{alg:mackay_algorithm}
	\begin{algorithmic}[1]
		\STATE {\bfseries Input:} Extracted features $F \in \mathbb{R}^{n\times D}$ and corresponding labels $y \in \mathbb{R}^n$ \\
		\STATE {\bfseries Output:} Logarithm of Maximum Evidence (LogME) \\
    \STATE {\bfseries Note:} $F$ has been pre-decomposed into $F = U\Sigma V^T$ \\ 

    \STATE{Initialize} $\alpha=1, \beta=1$ \\ 
      \WHILE{$\alpha, \beta$ not converge}
        \STATE{Compute} $\gamma = \sum_{i=1}^{D} \frac{\beta \sigma_i^2}{\alpha + \beta \sigma_i^2}, \Lambda = \text{diag}\{(\alpha + \beta \sigma^2)\}$ \\
        \STATE{\textbf{Na\"ive}}: $A = \alpha I + \beta F^TF, m = \beta A^{-1}F^Ty$ \\
        \STATE{\textbf{Optimized} by \citet{you_logme:_2021}}: $m = \beta (V (\Lambda^{-1} (V^T (F^T y))))$ \\
        \STATE{Update} $\alpha \leftarrow \frac{\gamma}{m^Tm}, \beta \leftarrow \frac{n - \gamma}{\vert \vert F m - y \vert \vert_2^2}$  \\
      \ENDWHILE
      \STATE {Compute} and return $\mathcal{L} = \frac{1}{n} \mathcal{L}(\alpha, \beta)$ using Equation~\ref{eq:evidence} 
		
	\end{algorithmic}

  \kern2pt\hrule\relax

  \bigskip
  \bigskip

	\algcaption{Evidence Maximization by Optimized Fixed Point Iteration}
	\label{alg:optimized_fixed_point}
	\begin{algorithmic}[1]
		\STATE {\bfseries Input:} Extracted features $F \in \mathbb{R}^{n\times D}$ and corresponding labels $y \in \mathbb{R}^n$ \\
		\STATE {\bfseries Output:} Logarithm of Maximum Evidence (LogME) \\
    \STATE {\bfseries Require:} Truncated SVD of $F$: $F = U_r\Sigma_r V_r^T$, with $U_r \in \mathbb{R}^{n\times r}, \Sigma_r \in \mathbb{R}^{r\times r}, V_r \in \mathbb{R}^{D\times r}$. \\ 
    \STATE {\bfseries Compute} the first $r$ entries of $z = U_{r}^T y$ \\
    \STATE {\bfseries Compute} the sum of remaining entries $\Delta = \sum_{i=r+1}^{n} z_i^2 = \sum_{i=1}^{n} y_i^2 - \sum_{i=1}^{r} z_i^2$ \\ 

    \STATE{Initialize} $\alpha=1, \beta=1, t = \frac{\alpha}{\beta} = 1$ \\ 
      \WHILE{$t$ not converge}
        \STATE{Compute} $m^Tm = \sum_{i=1}^{r} \frac{\sigma_i^2 z_i^2}{(t + \sigma_i^2)^2}$, $\gamma = \sum_{i=1}^{r} \frac{\sigma_i^2}{t + \sigma_i^2}$, $\vert \vert F m - y \vert \vert_2^2 = \sum_{i=1}^{r} \frac{z_i^2}{(1 + \sigma_i^2 / t)^2} + \Delta$\\
        \STATE{Update} $\alpha \leftarrow \frac{\gamma}{m^Tm}, \beta \leftarrow \frac{n - \gamma}{\vert \vert F m - y \vert \vert_2^2}, t = \frac{\alpha}{\beta}$  \\
      \ENDWHILE
      \STATE {Compute} $m = V_r \Sigma' z$, where $\Sigma'_{ii} = \frac{\sigma_i}{t + \sigma_i^2} (1 \le i \le r)$. 
      \STATE {Compute} and return $\mathcal{L} = \frac{1}{n} \mathcal{L}(\alpha, \beta)$ using Equation~\ref{eq:evidence} 		
	\end{algorithmic}
  \kern2pt\hrule\relax
\end{figure}

\subsubsection{Computational speedup}
\label{sec:speedup}

Although the Bayesian approach of maximum evidence has a rigorous theoretical explanation~\citep{knuth_bayesian_2015}, it inherits the drawback of Bayesian methods with respect to high computational complexity. A na\"ive implementation of Algorithm~\ref{alg:mackay_algorithm} results in a total complexity of $\mathcal{O}(CD^3 + nCD^2)$. For typical usage with $D \approx 10^3, n \approx 10^4, C \approx 10^3$, the computational cost is $10^{13}$, with the wall-clock time comparable to fine-tuning PTM $\phi$.

Our conference paper~\citep{you_logme:_2021} accelerated the computation by avoiding matrix inversion and matrix-matrix multiplication, as shown in Line~$8$ of Algorithm~\ref{alg:mackay_algorithm}. In this paper, we present a convergence analysis of MacKay's algorithm by fixed point iteration. It turns out that the analysis implies a faster algorithm for evidence maximization. The algorithm is presented in Algorithm~\ref{alg:optimized_fixed_point} and its rationale is explained in Section~\ref{sec:convergence_analysis}.

\begin{table}[htbp]
  \centering
  \caption{The complexity of Algorithm~\ref{alg:logme} with three implementations of evidence maximization. $n, C$ are the number of samples and the number of classes in classification (or the number of target variables in regression) in downstream tasks, and $D$ is the dimension of features produced by a pre-trained model.}
  \resizebox{.95\columnwidth}{!}{
    \begin{tabular}{ccc}
      \toprule
       Evidence maximization method  & Complexity per while-loop & Overall complexity \\
          \midrule
    na\"ive implementation &   $\mathcal{O}(D^3 + n D^2)$    & $\mathcal{O}(nCD^2 + CD^3)$ \\
    optimized by \citet{you_logme:_2021} &   $\mathcal{O}(D^2 + n D)$    & $\mathcal{O}(n D^2 + nCD + CD^2 + D^3)$ \\
    fixed point iteration (this paper) &   $\mathcal{O}(n)$    & $\mathcal{O}(nD^2 + nCD)$ \\
    \bottomrule
    \end{tabular}}
  \label{tab:optimization}%
\end{table}%

Table~\ref{tab:optimization} compares the complexity of calculating LogME with three implementations of evidence maximization. The na\"ive implementation is biquadratic, \citet{you_logme:_2021} made it cubic, and this paper further reduces the number of cubic terms. The optimized algorithm makes a time-consuming Bayesian approach fast enough, reducing the wall-clock time by order of $10^2$ (see Section~\ref{sec:timecost} for a quantitative measurement). Note that three implementation methods are functionally equivalent and only differ in computational complexity. Therefore, \emph{the fixed point iteration proposed in this paper is used by default  in our implementation}.

\section{Theoretical Analyses of LogME}
\label{sec:theoretical}

In this section, we analyze two theoretical aspects of the proposed LogME, which further explains the rationale behind the LogME algorithm and helps provide insight into why LogME works.

\subsection{Convergence analysis of evidence maximization}
\label{sec:convergence_analysis}

\textbf{Historical remarks:} The evidence maximization procedure in Section~\ref{sec:evidence_maximization} was proposed by \citet{mackay_bayesian_1992} as a heuristic method to maximize the evidence of given data, following the spirit of empirical Bayesian learning~\citep{bishop_neural_1995}. Theoretical analysis is missing and it has been employed as a heuristic in modern machine learning practice. Progress was made in the theoretical justification by~\citet{li_hyper-parameter_2016} who noted that if the predictive uncertainty $\beta$ is known, the maximization over model uncertainty $\alpha$ can be viewed as a special instantiation of the EM algorithm~\citep{dempster_maximum_1977}. However, pre-determining $\beta$ is suboptimal, and in practice $\alpha, \beta$ are simultaneously maximized. In this paper we provide an analysis of  MacKay's algorithm in which $\alpha, \beta$ are jointly optimized.

We collect necessary notation here: $n$ is the number of data examples; $D$ is the size of feature dimensionality; $F \in \mathbb{R}^{n\times D}$ is the feature matrix, with $r = \text{rank}(F)$ being its rank; $y \in \mathbb{R}^{n}$ is the label vector of data examples. We have that $r \le \min \{n, D\}$.

The key in our analysis is to take full advantage of the singular value decomposition of the feature matrix $F = U \Sigma V^T$, where $U \in \mathbb{R}^{n\times n}$, $V \in \mathbb{R}^{D\times D}$, and $\Sigma \in \mathbb{R}^{n\times D}$. Note that $\Sigma$ only has $r$ non-zero entries: $\Sigma_{ii} = \sigma_i > 0 \; (1 \le i \le r)$ where $\sigma_i^2$ is the $i$-th largest eigenvalue of $F^TF$ and $\sigma_i=0\ (r+1 \le i \le \max(n, D))$. To simplify the expression, let $z = U^Ty$ be the transformed $y$ under orthogonal bases $U$, \emph{i.e.}, $y = Uz$.

MacKay's algorithm (Algorithm~\ref{alg:mackay_algorithm}) consists of a while-loop which is presented in Algorithm~\ref{alg:iteration}.  The key to analyzing the whole algorithm is to analyze each iteration of the while-loop. During each iteration, new values $\alpha', \beta'$ are computed based on old values $\alpha, \beta$, which can be regarded as evaluating a vector-valued function $(\alpha', \beta') =g(\alpha, \beta)$.

\begin{algorithm}[htbp]
	\caption{One iteration of evidence maximization in Algorithm~\ref{alg:mackay_algorithm}.}
	\label{alg:iteration}
	\begin{algorithmic}[1]
		\STATE { Input:} $\alpha, \beta$; { Output:} $\alpha', \beta'$ for the next iteration.

        \STATE{Compute} $A = \alpha I + \beta F^TF, m = \beta A^{-1}F^Ty, \gamma = \sum_{i=1}^{D} \frac{\beta \sigma_i^2}{\alpha + \beta \sigma_i^2}$ \\
        \STATE{Return} $\alpha' = \frac{\gamma}{m^Tm}, \beta' = \frac{n - \gamma}{\vert \vert F m - y \vert \vert_2^2}$  \\		
	\end{algorithmic}
\end{algorithm}

MacKay's algorithm converges if and only if $(\alpha', \beta') = (\alpha, \beta)$ in Algorithm~\ref{alg:iteration}. With $F, y$ as constants, the convergence of Algorithm~\ref{alg:mackay_algorithm} is equivalent to the existence of the fixed point of the vector-valued function $g$, \emph{i.e.}, the existence of $\alpha, \beta $ such that $ (\alpha, \beta) =g(\alpha, \beta)$.

In general, fixed points of vector-valued functions are difficult to analyze and visualize. Fortunately, we find that the vector-valued function $(\alpha', \beta') =g(\alpha, \beta)$ is homogeneous: $g(k\alpha, k\beta) = k g(\alpha, \beta), \forall k > 0$. Let $t = \alpha / \beta$, and $t' = \alpha' / \beta'$, the vector-valued function $(\alpha', \beta') =g(\alpha, \beta)$ induces a scalar function $t' = f(t)$, whose explicit form can be derived in Theorem~\ref{thm:f}. Evaluating $g(\alpha, \beta)$ is equivalent to calculating $f(\frac{\alpha}{\beta})$, which is easier to analyze.

\begin{theorem}
  \textup{Algorithm~\ref{alg:iteration} induces a scalar function (Equation~\ref{eq:fixpoint})} \textup{with} $t = \frac{\alpha}{\beta}$ \textup{and} $t' = \frac{\alpha'}{\beta'}$.
  \begin{equation}
    t' = f(t) = \left(\frac{n}{n - \sum_{i=1}^{D} \frac{ \sigma_i^2}{t + \sigma_i^2}} - 1\right) t^2 \frac{\sum_{i=1}^{n} \frac{z_i^2}{(t + \sigma_i^2)^2}}{\sum_{i=1}^{n} \frac{\sigma_i^2 z_i^2}{(t + \sigma_i^2)^2}}.
    \label{eq:fixpoint}
  \end{equation}
  \label{thm:f}
  \vspace{-10pt}
\end{theorem}

The proof is in Appendix~\ref{proof:f}. Although $f(t)$ seems very complicated and completely understanding its behavior is difficult, surprisingly, the existence of a fixed point of $f(t)$ can be guaranteed with an interpretable condition, as presented in the following Theorem~\ref{thm:converge}.

\begin{theorem}
  \textup{If} $r < n$ \textup{and} $\sum_{1 \le i,j \le n}(z_i^2-z_j^2)(\sigma_i^2-\sigma_j^2)>0$, \textup{then} $f(t)$ \textup{has a fixed point and thus MacKay's algorithm will converge.}
  \label{thm:converge}
\end{theorem}

\begin{figure}[htbp]
  \centering
  \includegraphics[width=\columnwidth]{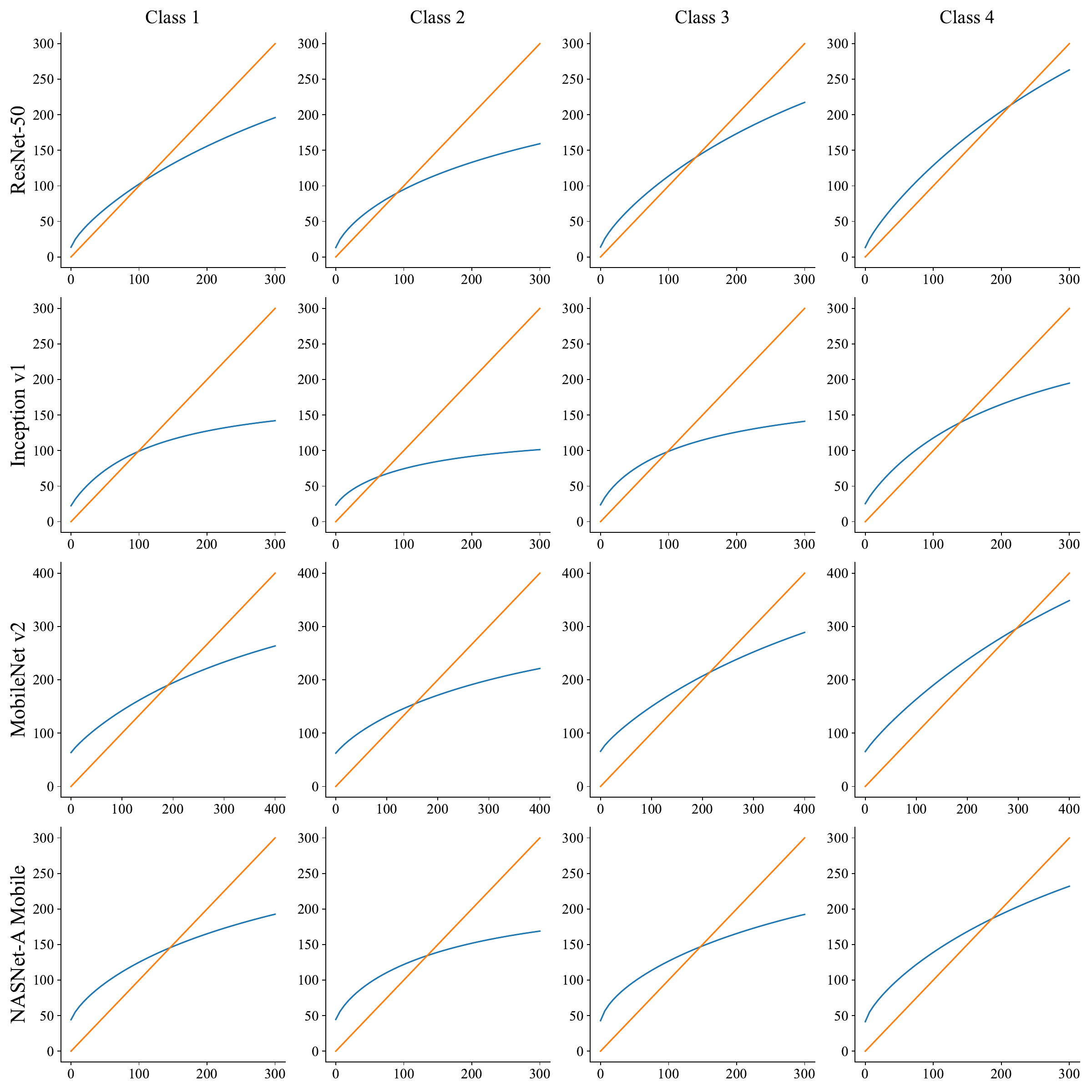}
  \includegraphics[width=.2\columnwidth]{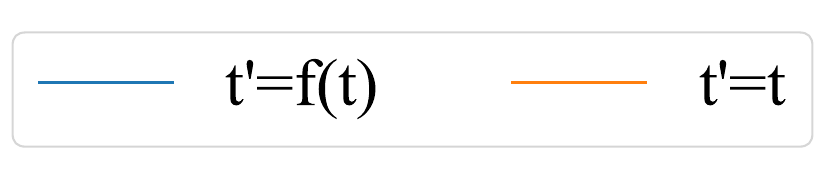}
  \caption{Fixed points of $f(t)$ in Equation~\ref{eq:fixpoint} for the first four classes in CIFAR10 with four pre-trained models. The full figure for all the ten classes in CIFAR10 with five pre-trained models can be found in Figure~\ref{fig:full}, which is omitted here to prettify the layout. We plot $t' = f(t)$ (in blue) and $  t' = t$ (in orange), whose intersections are fixed points $f(t) = t$. The existence of fixed points guarantees the convergence of MacKay's algorithm for evidence maximization.}
  \label{fig:fixpoint}
\end{figure}

The proof is in Appendix~\ref{proof:converge}. Theorem~\ref{thm:converge} requires two conditions to guarantee the fixed point: $r < n$ and $\sum_{1 \le i,j \le n}(z_i^2-z_j^2)(\sigma_i^2-\sigma_j^2)>0$. The first condition is easy to interpret and can be easily satisfied: usually $n > D$, and $n > D \ge r$ naturally holds. The condition $\sum_{1 \le i,j \le n}(z_i^2-z_j^2)(\sigma_i^2-\sigma_j^2)>0$ is new in this paper. Note that $z = U^Ty$ and $z_i = U_i^Ty$, where $U_i$ (the $i$-th column of $U$) is the left-singular vector of the singular value $\sigma_i$, which means that $z_i$ is the projection of label vector $y$ in the direction of the left-singular vector for the singular value $\sigma_i$. Intuitively speaking, $\sum_{1 \le i,j \le n}(z_i^2-z_j^2)(\sigma_i^2-\sigma_j^2)>0$ requires $z_i^2$ to share roughly the same descending order as $\sigma_i^2$. For larger $\sigma_i^2$ (\emph{i.e.}, smaller $i$), it means the projection of $y$ in the corresponding left-singular vector should be larger, which can be interpreted as a rigorous way to say that \textit{labels $y$ are meaningful with respect to the features $F$}. We would like to emphasize that the requirement on the order of $z_i^2$ is \textit{soft}: strict order $z_i^2 \ge z_j^2 \iff i \le j \iff \sigma_i^2 \ge \sigma_j^2$ certainly assures the convergence condition $\sum_{1 \le i,j \le n}(z_i^2-z_j^2)(\sigma_i^2-\sigma_j^2)>0$, but as long as most $z_i^2$ follow the order, the condition can be satisfied. We find that all experiments in this paper admit the convergence condition, \emph{i.e.}, the evidence maximization algorithm is guaranteed to converge if the data is meaningful. For example, Figure~\ref{fig:fixpoint} plots $f(t)$ on the CIFAR10 dataset, which clearly shows cross points of $f(t)$ and $t$, so the convergence condition $\sum_{1 \le i,j \le n}(z_i^2-z_j^2)(\sigma_i^2-\sigma_j^2)>0$ holds.

\textbf{Make the fixed point iteration faster}. Note that the fixed point iteration Equation~\ref{eq:fixpoint} requires explicitly computing $z = U^Ty$ with $\mathcal{O}(n^2)$ storage and computation, which would be undesirable if $n$ is very large. To obtain a practical algorithm, we take advantage of the fact that $\sigma_i = 0$ for $i > r$, and optimize the fixed point iteration as follows:
\begin{align*}
  t' = f(t) &= \left(\frac{n}{n - \sum_{i=1}^{D} \frac{ \sigma_i^2}{t + \sigma_i^2}} - 1\right) t^2 \frac{\sum_{i=1}^{n} \frac{z_i^2}{(t + \sigma_i^2)^2}}{\sum_{i=1}^{n} \frac{\sigma_i^2 z_i^2}{(t + \sigma_i^2)^2}} \\
  &= \left(\frac{n}{n - \sum_{i=1}^{r} \frac{ \sigma_i^2}{t + \sigma_i^2}} - 1\right) t^2 \frac{\sum_{i=1}^{r} \frac{z_i^2}{(t + \sigma_i^2)^2} + \frac{1}{t^2} \sum_{i=r+1}^{n} z_i^2}{\sum_{i=1}^{r} \frac{\sigma_i^2 z_i^2}{(t + \sigma_i^2)^2}} \\
  &= \left(\frac{n}{n - \sum_{i=1}^{r} \frac{ \sigma_i^2}{t + \sigma_i^2}} - 1\right) t^2 \frac{\sum_{i=1}^{r} \frac{z_i^2}{(t + \sigma_i^2)^2} + \frac{1}{t^2} (\sum_{i=1}^{n} y_i^2 - \sum_{i=1}^{r} z_i^2)}{\sum_{i=1}^{r} \frac{\sigma_i^2 z_i^2}{(t + \sigma_i^2)^2}}.
\end{align*}
Therefore, we can derive a faster algorithm (Equation~\ref{eq:final_fixedpoint}) for the fixed point iteration, which only requires the first $r$ entries of $z$ without computing the full $U$ matrix or the full $z$ vector. This is the algorithm we implement in Algorithm~\ref{alg:optimized_fixed_point}, setting:
\begin{equation}
  t' = f(t) = \left(\frac{n}{n - \sum_{i=1}^{r} \frac{ \sigma_i^2}{t + \sigma_i^2}} - 1\right) t^2 \frac{\sum_{i=1}^{r} \frac{z_i^2}{(t + \sigma_i^2)^2} + \frac{1}{t^2} (\sum_{i=1}^{n} y_i^2 - \sum_{i=1}^{r} z_i^2)}{\sum_{i=1}^{r} \frac{\sigma_i^2 z_i^2}{(t + \sigma_i^2)^2}}.
  \label{eq:final_fixedpoint}
\end{equation}

\subsection{Influence of dimensionality}
\label{sec:influence_of_dimension}

In Section~\ref{sec:evidence_maximization}, we normalize the LogME value by the number of examples because Equation~\ref{eq:evidence} scales linearly with $n$. The influence of feature dimension $D$ is, however, unclear. In this section, we find two cases (feature duplicate and feature padding) where \textit{LogME value remains unchanged when the feature dimension goes up without introducing more information.} The two cases show the existence of infinitely many features with arbitrary feature dimensions that share the same LogME value, therefore removing the necessity of dimensionality normalization.

\begin{corollary}[feature duplicate]
  \textup{LogME value will remain the same if the feature consists of arbitrary replicas of the original feature. Formally speaking, if the LogME value for }$F \in \mathbb{R}^{n \times D}$ \textup{and} $y \in \mathbb{R}^{n}$ \textup{is} $\mathcal{L},$ \textup{then the LogME value for} $\tilde{F}=[F,..., F]\in R^{n\times qD}$ \textup{and} $y \in \mathbb{R}^{n}$ \textup{is also} $\mathcal{L}. \; (q \in \mathbb{N}$ \textup{is a natural number to represent the number of replicas.)} 
  \label{thm:duplicate}
\end{corollary}

\begin{corollary}[feature padding]
  \textup{LogME value will remain the same if the feature is padded with an arbitrary number of zeros. Formally speaking, if the LogME value for }$F \in \mathbb{R}^{n \times D}$ \textup{and} $y \in \mathbb{R}^{n}$ \textup{is} $\mathcal{L},$ \textup{then the LogME value for} $\tilde{F}=[F, \mathbf{0}]\in R^{n\times (D+d)}$ \textup{and} $y \in \mathbb{R}^{n}$ \textup{is also} $\mathcal{L}. \; (d \in \mathbb{N}$ \textup{is a natural number and} $\mathbf{0}\in \mathbb{R}^{n\times d}$ \textup{is a matrix with all zero entries.)}
  \label{thm:padding}
\end{corollary}

The proofs of Corollary~\ref{thm:duplicate} and Corollary~\ref{thm:padding} are in Appendix~\ref{proof:duplicate} and Appendix~\ref{proof:padding}, respectively. The core idea is to find the closed-form relationship between decompositions of $\tilde{F}$ and $F$.

Corollary~\ref{thm:duplicate} and Corollary~\ref{thm:padding} imply that duplicating features or padding features with zeros will not change the value of LogME. LogME is capable of filtering out redundant information in features, explaining its excellent empirical performance in \citet{you_logme:_2021}.

\section{Tuning Pre-Trained Models}
\label{sec:tuning_ptms}

The new paradigm we propose consists of ranking and tuning pre-trained models. Sections~\ref{sec:ranking_ptms} and \ref{sec:theoretical} described technical background for ranking pre-trained models, including the transferability metric LogME and its theoretical analyses. In this section we focus on tuning pre-trained models, completing the overall paradigm.

We identify two possible scenarios in the tuning of pre-trained models: single best PTM tuning and multiple PTM tuning. (1) \emph{Single best PTM tuning} is suitable if there are no constraints on the network architecture, parameter count, or FLOPs of computation. These constraints are common in industrial applications, but are less important in academic research. Therefore, single best PTM tuning is common in academic research. Intuitively, the remaining PTMs are considered inferior to the best ranked PTM so they are not considered worth the effort to identify and deploy. We refer readers to dedicated papers~\citep{chen_catastrophic_2019,kou_stochastic_2020,you_co-tuning_2020} on how to fine-tune a single PTM. (2) When we deploy neural networks in industrial applications, typically there are strict constraints on the budget of memory footprint or power consumption. Therefore, the pre-trained model $\phi_t$ satisfying these constraints is probably not the best ranked. The current state of the art is that practitioners can only fine-tune $\phi_t$, and the overall knowledge of the PTM hub $\{\phi_{i}\}_{i=1}^M$ cannot be exploited. In this paper, we show that it is possible to transfer knowledge from several teacher PTMs $\{\phi_{k}\}_{k=1}^K$ to the target pre-trained model $\phi_t$ during fine-tuning, a paradigm which we call ``multiple PTM tuning.''

A side issue in tuning multiple PTMs is how to select the teacher PTMs. Typically, $K < M$, \emph{i.e.}, not all PTMs are necessary, since some PTMs may not be suitable for the target task and would hinder the transfer learning procedure. However, for $M$ pre-trained models, the possible number of teacher combinations is $\mathcal{O}(2^M)$, which is impractical to enumerate. To overcome the exponential complexity, we avail ourselves of the PTM ranking. With the PTM ranking, we can greedily select teacher PTMs according to the rank. For example, if we want to choose $K$ teacher PTMs, then the top-$K$ ranked PTMs $\{\phi_{k}\}_{k=1}^K$ are the teacher for knowledge transfer. As for how to choose the hyper-parameter $K \; (1 \le K \le M)$, we give empirical guidelines in Section~\ref{sec:exp_tuning_ptms}.

Multiple PTM tuning offers a unique advantage over simply fine-tuning the target pre-trained model $\phi_t$: if the specified target pre-trained model $\phi_t$ is not the best-ranked, we can still improve it by transferring knowledge from top-performing PTMs $\{\phi_{k}\}_{k=1}^K$.

\subsection{Problem setup for multiple PTM tuning}
\label{sec:setup_multiple_ptms_tuning}

Now suppose we have selected $K$ pre-trained models $\{\phi_{k}\}_{k=1}^K$, with each pre-trained model $\phi_{k}$ transforming input $x$ into a $D_k$-dimensional feature vector. In general, pre-trained models $\{\phi_{k}\}_{k=1}^K$ have various network architectures, and dimensionality of features $\{D_{k}\}_{k=1}^K$ can vary. Let $\phi_t$ be the target architecture, which transforms the input into $D_t$ dimensional feature vector. Formally, the multiple PTM tuning problem is to fine-tune a pre-trained model $\phi_t$ by leveraging selected pre-trained models $\{\phi_{k}\}_{k=1}^K$, as shown in Figure~\ref{fig:overview}.

To fine-tune a model $\phi_t$ in a target task, a new output head would be attached after $\phi_t$, where a target-specific loss is calculated. The target-specific head and loss are necessary for every possible solution to multiple PTM tuning, which is taken care of by a loss function $L_{task}$. We will not elaborate on $L_{task}$ as it varies with tasks. Next, in Section~\ref{sec:existing_method_multiple}, we summarize existing approaches to the problem and introduce our method in Section~\ref{sec:bayesian_tuning}.

\subsection{Existing approaches to the problem of multiple PTM tuning}
\label{sec:existing_method_multiple}

\textbf{A baseline approach} is to fine-tune $\phi_t$ without considering teacher PTMs $\{\phi_{k}\}_{k=1}^K$. This can serve as a baseline to measure the improvement brought by multiple PTM tuning.

\textbf{The knowledge distillation
approach to multiple PTM tuning} is knowledge distillation~\citep{hinton2015distilling} in the feature space via mean-square error. Since the feature dimensions may differ between $\phi_t$ and $\{\phi_{k}\}_{k=1}^K$, a transformation module is necessary. The knowledge distillation (KD) method takes advantage of selected pre-trained models by adding a regularization term. $L_{KD} = \frac{1}{n} \sum_{i=1}^n \frac{1}{K} \sum_{k=1}^K \vert \vert \phi_k (x_i) - W_k \phi_t(x_i) \vert \vert_2^2$, where $W_k$ is learnable parameter to transform a $D_t$-dimensional feature $\phi_t(x_i)$ into a $D_k$-dimensional vector compatible with $\phi_k (x_i)$. Even if $D_k = D_t$, the semantics of each dimension in $\phi_t$ and $\phi_k$ may vary, making it necessary to introduce the transformation parameter $W_k$. The final loss is $L_{task} + \lambda L_{KD}$, with hyper-parameter $\lambda$ trading the two terms. The KD method is another simple but general baseline in multiple PTM tuning. It can be applied to various PTMs but the performance improvement is limited.

\textbf{Zoo-tuning for homogeneous PTMs tuning.} In the special case when $\phi_t$ and $\{\phi_{k}\}_{k=1}^K$ all share the same network architecture, Zoo-tuning proposed by \citet{shu_zoo-tuning:_2021} adaptively aggregates parameters of $\{\phi_{k}\}_{k=1}^K$ into $\phi_t$ in a layer-wise fashion. It does not modify the loss $L_{task}$, but changes the training process by model aggregation. Zoo-tuning is the current state-of-the-art method for homogeneous PTM tuning, but it fails to deal with the heterogeneous scenario when architectures of $\phi_t$ and $\{\phi_{k}\}_{k=1}^K$ are different.

\subsection{B-Tuning: A Bayesian approach to multiple PTM tuning}
\label{sec:bayesian_tuning}

We draw lessons from the shortcomings of the aforementioned knowledge distillation approach and the Zoo-tuning approach. Knowledge distillation operates at the level of output features, which works for heterogeneous PTMs but aligning features across PTMs is not easy. Zoo-tuning operates at the level of parameters (layers), thereby limiting itself to the homogeneous case. Taking the positive aspects of both frameworks, we design our approach to operate at the level of features to hide the heterogeneity among PTMs, and we go beyond features to avoid explicitly aligning features from various pre-trained models. Inspired by the ranking metric (LogME), we propose an approach that builds on posterior predictive distributions from Bayesian regression.

\textbf{A posterior predictive distribution} is $p(y' | f, F, y) = \int_{w} p(y' | w, f) p(w | F, y) {\rm d} w $, which predicts the label $y'$ of incoming feature $f$ conditioned on all the available training features $F$ and labels $y$ rather than just using $f$. With pre-computed $\alpha^*, \beta^*, m$ (byproducts of the LogME algorithm), $p(y' | w, f) \sim \mathcal{N}(w^Tf, {\beta^*}^{-1})$ by definition, and $p(w | F, y) = \frac{p(w) p(F, y | w)}{\int_{w'} p(w') p(F, y | w') {\rm d} w'}$ by Bayes' theorem. \citet{rasmussen_gaussian_2003} shows that $p(w | F, y) \sim \mathcal{N}(\beta^* A^{-1}F^Ty, A^{-1})$ with $A = \alpha^* I + \beta^* F^TF$. Plugging in the distributions of $p(y' | w, f)$ and $p(w | F, y)$, \citet{rasmussen_gaussian_2003} shows that $p(y' | f, F, y) \sim \mathcal{N}(f^Tm, f^TA^{-1}f + {\beta^*}^{-1})$ with $m = \beta^* A^{-1}F^Ty$. In short, for extracted features $F \in \mathbb{R}^{n\times D}$ and labels $y \in \mathbb{R}^n$, the LogME algorithm gives $\alpha^*, \beta^*, m$, and the posterior predictive distribution is $p(y' | f, F, y) \sim \mathcal{N}(f^Tm, f^TA^{-1}f + {\beta^*}^{-1})$. For a full derivation of the posterior predictive distribution please see \citet{rasmussen_gaussian_2003}.

\begin{figure}[htbp]
  \centering
  \includegraphics[width=\columnwidth]{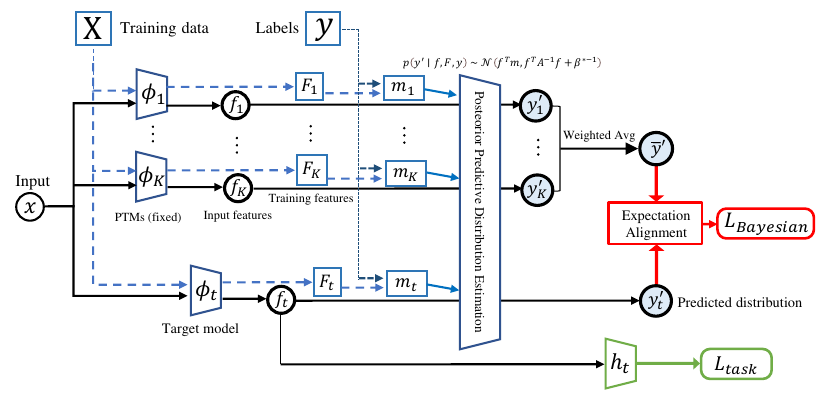}
  \vspace{-20pt}
  \caption{Illustration of B-Tuning. Dashed lines are pre-calculated before tuning.}
  \label{fig:btuning}
\end{figure}

The posterior predictive distribution depends on the training data ($F, y$) and input features $f$. Let $F_{k}$ be features extracted by the pre-trained model $\phi_k$, $f_{k} = \phi_k(x)$ be the output features of the current data point extracted by the pre-trained model $\phi_k$, then each pre-trained model can produce a posterior predictive distribution $p(y_{k}' | f_k, F_k, y) \sim \mathcal{N}(f_k^Tm_{k}, f_k^TA_{k}^{-1}f_k + {\beta_k^*}^{-1})$. How to combine these distributions? We propose to mix them according to their LogME values $\{\mathcal{L}_{k}\}_{k=1}^K$ as a mixture of Gaussians, $\bar{y}' = \sum_{k=1}^K \pi_k y_{k}'$ where $\pi_k = \frac{\exp (\mathcal{L}_{k} / t)}{\sum_{j=1}^K \exp (\mathcal{L}_{j} / t)}$, where $t$ is a temperature hyper-parameter~\citep{hinton2015distilling} that can be adjusted according to the difference of LogME values. Although $\{y_{k}'\}_{k=1}^K$ admit simple Gaussian distributions, the exact distribution of mixed $\bar{y}'$ is intractable because the features $F_{k}$ come from the same dataset and $\{y_{k}'\}_{k=1}^K$ are dependent. Nevertheless, according to the linearity property of expectation, $\mathbb{E} \bar{y}' = \sum_{k=1}^K \pi_k \mathbb{E} y_{k}' = \sum_{k=1}^K \pi_k f_k^T m_{k}$, \emph{the expectation of $\bar{y}'$ is known.}

For the target model $\phi_t$, the posterior predictive distribution is defined as $p(y_{t}' | f_t, F_{t}, y) \sim \mathcal{N}(f_t^Tm_{t}, f_t^TA_{t}^{-1}f_t + {\beta_t^*}^{-1})$. Since $\bar{y}'$ can be regarded as prior knowledge from pre-trained models $\{\phi_{k}\}_{k=1}^K$, we can \emph{align the expectation of $y_{t}'$ and $\bar{y}'$} as a regularization term, $L_{Bayesian} = \frac{1}{n}\sum_{i=1}^n \vert \vert \mathbb{E} \bar{y}' - \mathbb{E} y'_t \vert \vert_2^2$. Note that the expectation is taken over the predictive distributions and can be calculated analytically. Extending the formula to multiple classes, the final expression of the regularization term is
\begin{equation}\label{LBayesian}
  L_{Bayesian} = \frac{1}{n}\sum_{i=1}^n \frac{1}{C} \sum_{c=1}^C (\sum_{k=1}^K \pi_k f_k^Tm_{k, c} - f_t^Tm_{t, c})^2, 
\end{equation}
where $m_{k, c}, m_{t, c}$ are calculated by the LogME algorithm and are fixed during training. The final loss is $L_{task} + \lambda L_{Bayesian}$, with $\lambda$ introduced to trade off two terms. Because the method depends on the Bayesian approach of calculating posterior predictive distribution, we call it Bayesian Tuning, or \textbf{B-Tuning}. Figure~\ref{fig:btuning} describes the method and the computation graph. Only $\phi_t$ is updated during B-Tuning while the teacher PTMs $\{\phi_{k}\}_{k=1}^K$ are fixed.

B-Tuning has two advantages over previous methods: (1) It hides the heterogeneity among PTMs by operating at the level of features, yielding a \emph{general} solution to multiple PTM tuning for both the homogeneous and heterogeneous cases. (2) B-Tuning has a simple interpretation: it aligns features adaptively with $m$ serving as an attention-like mechanism, removing the necessity of learning to transform features into a shared space as in the knowledge distillation approach. The superiority of B-Tuning in multiple PTM tuning is empirically demonstrated in Section~\ref{sec:exp_tuning_ptms}.

\section{Experiments}
\label{sec:experiments}

This section presents comprehensive experiments. Section~\ref{sec:toy} illustrates the behavior of LogME on toy problems. Experiments on ranking PTMs and tuning PTMs are in Section~\ref{sec:exp_on_ranking_ptms} and Section~\ref{sec:exp_tuning_ptms} respectively, demonstrating the power of the proposed new paradigm. Section~\ref{sec:timecost} quantitatively measures the efficiency of LogME and Section~\ref{sec:compare_head} compares LogME against a common approach of re-training the head over a fixed feature extractor, providing a comprehensive understanding of LogME. Original data for some figures are available in the appendix. Code for LogME is available at \url{https://github.com/thuml/LogME}.

\subsection{Illustration with toy data}
\label{sec:toy}

To give an intuitive sense of how LogME works, we generate features with increasing noise to mimic the features extracted by pre-trained models with decreasing transferability and to check if LogME can measure the quality of features.

\begin{figure}[t]
  \centering
  \subfigure{
    \includegraphics[width=.45\columnwidth]{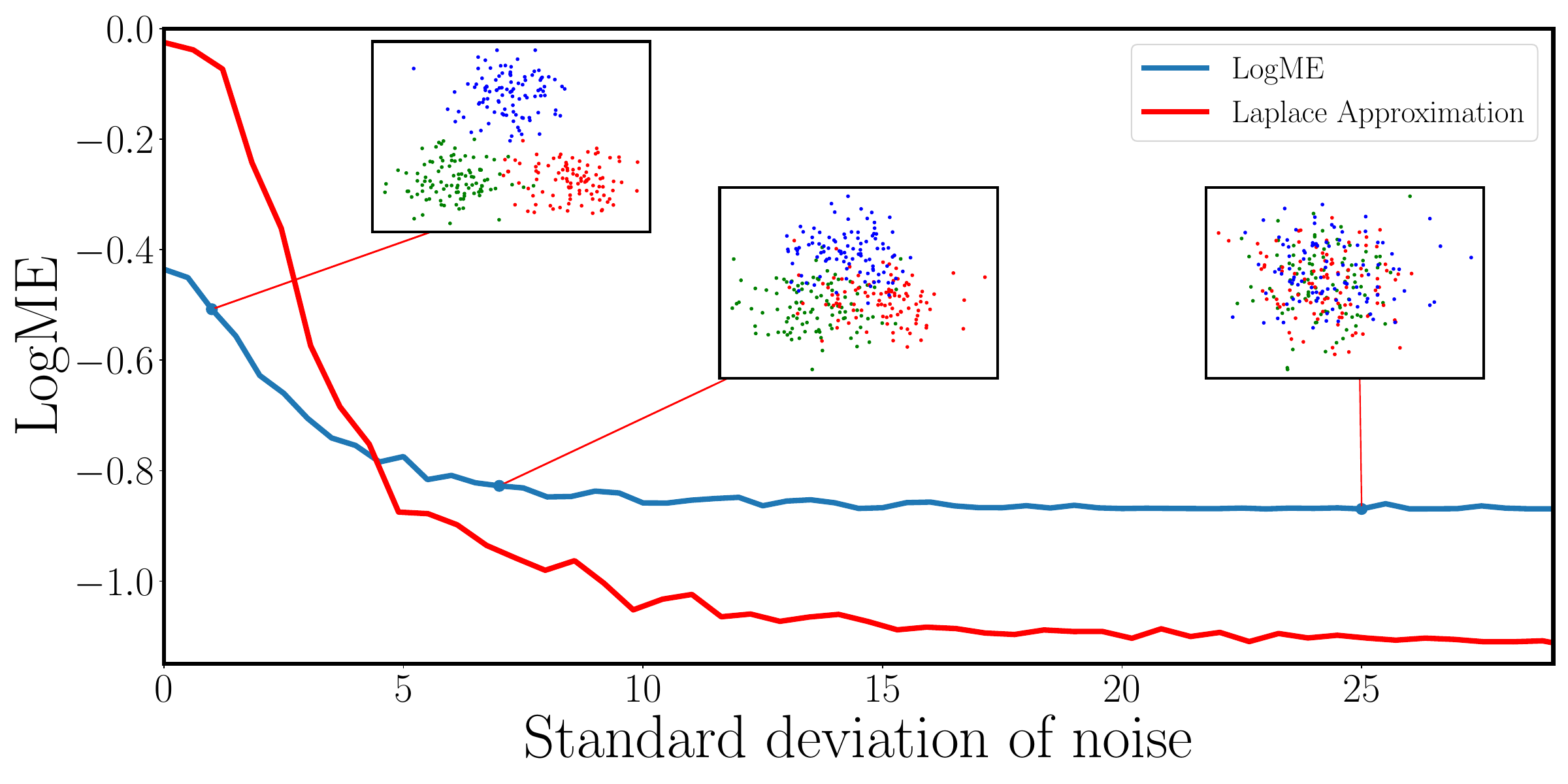}
  }
  \subfigure{
    \includegraphics[width=.45\columnwidth]{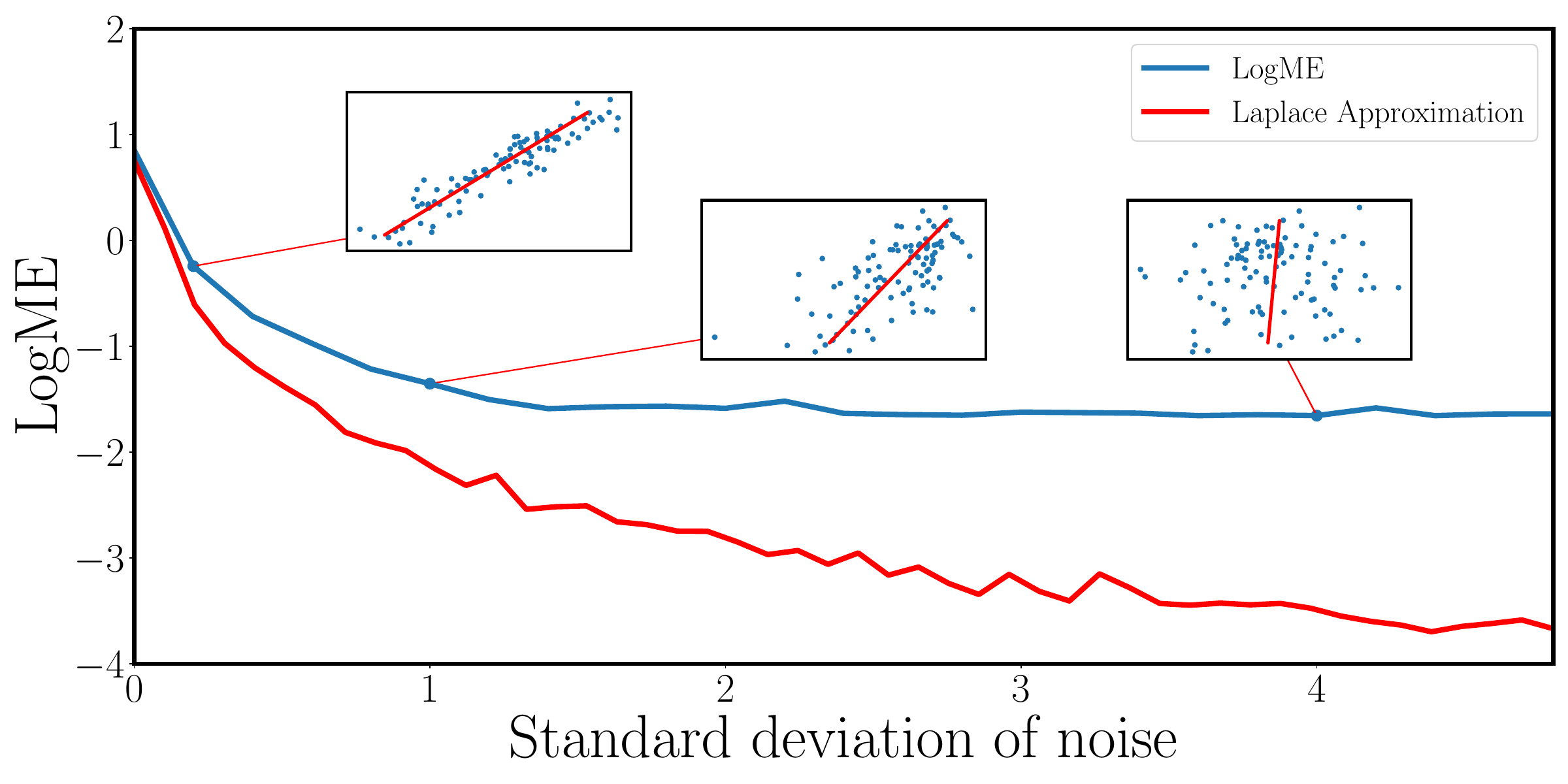}
  }
  \caption{Experiments with toy data demonstrate that LogME values go down with decreasing feature quality. ``Laplace Approximation'' means that the LogME value calculated by \citet{immer_scalable_2021} use the Laplace approximation.}
  \vspace{-10pt}
  \label{fig:toy}
\end{figure}

For classification (Figure~\ref{fig:toy} left), three clusters in a 2-D plane are generated, with colors representing the categories. Initially, the features are well separated so LogME has a large value. Then we add Gaussian noise with increasing variance to the data and the clustering structure in feature space disappears, leading to smaller LogME values as expected.

For regression (Figure~\ref{fig:toy} right), $x$ is uniformly distributed ($x \sim \mathcal{U}[0, 1]$) and the output $y = 2 x + \epsilon$ with observation error $\epsilon \sim \mathcal{N}(0, 0.1^2)$. By adding noise to the feature $x' = x + \mathcal{N}(0, t^2)$, the quality of feature $x'$ becomes worse and it is harder to predict $y$ from $x'$. With larger $t$ (the standard deviation of noise), LogME becomes smaller as expected.

These toy experiments on synthesized data show that LogME is an effective measure of the feature quality, and therefore can provide a ranking for PTMs in a pre-trained model hub.

Figure~\ref{fig:toy} also shows the LogME value calculated by \citet{immer_scalable_2021} using Laplace approximation. In this toy experiment, both LogME and Laplace approximation correctly measure the trend of feature quality. In regression, the Laplace Approximation is strictly lower than LogME; in classification, Laplace approximation uses a categorical prior and approximates the marginal likelihood, while LogME converts classification labels to one-hot labels (with a Gaussian prior) and calculates the exact value without approximation. The left plot in Figure~\ref{fig:toy} confirms that both approaches can reflect the trend of feature quality. However, we notice that Laplace approximation has larger fluctuations than LogME, and the Laplace approximation requires more computation than LogME. In addition, its performance in realistic data (Section~\ref{sec:supervised_classification}) is not satisfactory. Therefore, when dealing with classification data, we convert classification labels to one-hot labels and treat the problem as a multivariate regression problem in LogME. How to analytically calculate the value with a categorical prior is left as a future research question.

\subsection{Ranking pre-trained models}
\label{sec:exp_on_ranking_ptms}

This section focuses on the first part of the proposed paradigm: ranking pre-trained models. The goal is to rank pre-trained models so that potentially best PTMs can be selected for the subsequent tuning process. This section attaches great importance to the diversity of pre-trained models and downstream tasks. Section~\ref{sec:supervised_classification} and Section~\ref{sec:supervised_regression} transfer supervised pre-trained models to classification and regression tasks, respectively; Section~\ref{sec:constrastive_ptms} explores unsupervised pre-trained models on both classification and regression; Section~\ref{sec:language_ptm_to_glue} and Section~\ref{sec:ner_experiments} study pre-trained language models on language understanding tasks and a sequential tagging task, respectively. These extensive experiments demonstrate the generality and effectiveness of the proposed LogME method in ranking pre-trained models.

\subsubsection{Ranking supervised pre-trained models in classification tasks}
\label{sec:supervised_classification}

We use $12$ ImageNet pre-trained models available from PyTorch: Inception~V1~\citep{szegedy_going_2015}, Inception~V3~\citep{szegedy_rethinking_2016}, ResNet~34~\citep{he_deep_2016}, ResNet~50~\citep{he_deep_2016}, ResNet~101~\citep{he_deep_2016}, ResNet~152~\citep{he_deep_2016}, Wide ResNet~50~\citep{zagoruyko_wide_2017}, DenseNet~121~\citep{huang_densely_2017}, DenseNet~169~\citep{huang_densely_2017}, DenseNet~201~\citep{huang_densely_2017}, MobileNet~V2~\citep{sandler_mobilenetv2:_2018}, and NASNet-A Mobile~\citep{tan_mnasnet:_2019}. These pre-trained models cover most of the supervised pre-trained models in transfer learning that practitioners frequently use.

For downstream classification tasks, we take nine commonly used datasets: Aircraft~\citep{maji_fine-grained_2013}, Birdsnap~\citep{berg_birdsnap:_2014}, Caltech~\citep{fei-fei_learning_2004}, Cars~\citep{krause_collecting_2013}, CIFAR10~\citep{krizhevsky_learning_2009}, CIFAR100~\citep{krizhevsky_learning_2009}, DTD~\citep{cimpoi_describing_2014}, Pets~\citep{parkhi_cats_2012}, and SUN~\citep{xiao_sun_2010}. The description of each dataset and data statistics are listed in Appendix~\ref{app:dataset}.

\begin{figure*}[tbp]
  \centering
  \includegraphics[width=.8\textwidth]{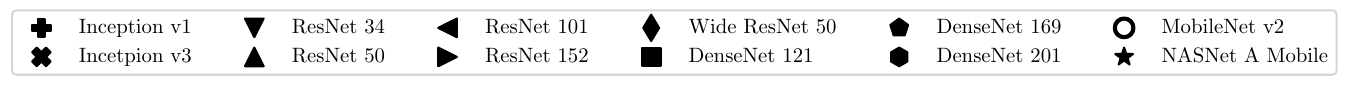}
  \begin{center}
  \includegraphics[width=\textwidth]{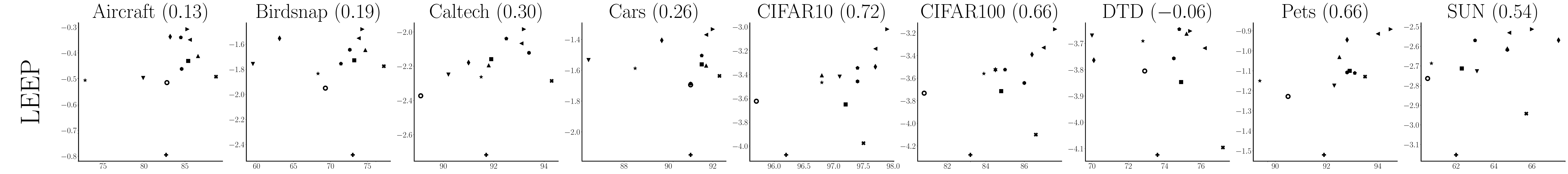}
  \end{center}
  \begin{center}
  \includegraphics[width=\textwidth]{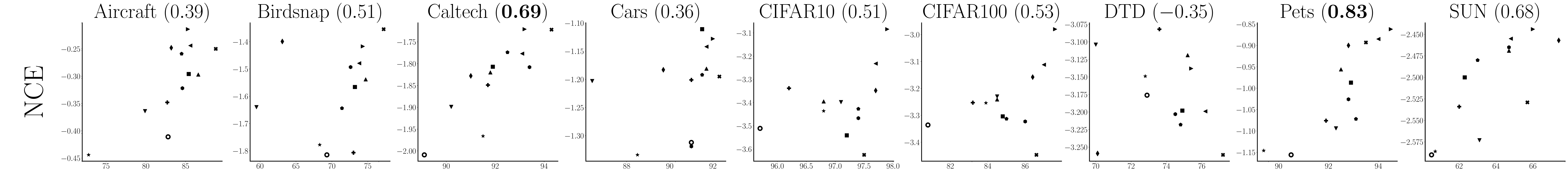}
  \end{center}
  \begin{center}
  \includegraphics[width=\textwidth]{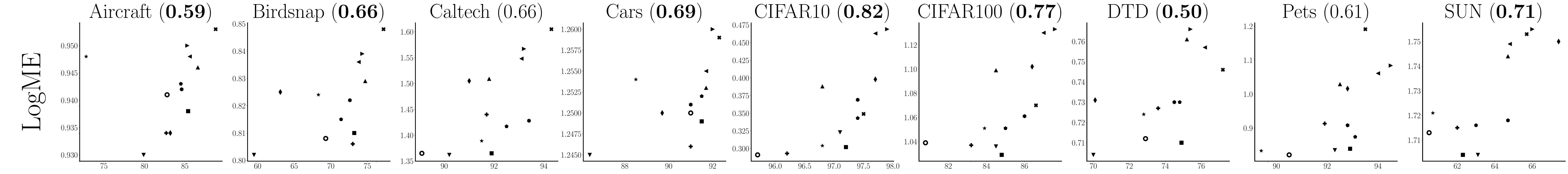}
  \end{center}
  \vspace{-10pt}
  \caption{Correlation values ($\tau_w$) between fine-tuned accuracy (X-axis) and scores produced by three methods (Y-axis) for ranking PTMs on 9 datasets with 12 pre-trained models. One row for each method, one column for each dataset (with $\tau_w$ in the parenthesis near the dataset name), and one marker for each pre-trained model. The best $\tau_w$ in each dataset is marked in bold.}
  \vspace{-10pt}
  \label{fig:pre_train_selection}
\end{figure*}

For all the datasets we use, we respect the official train/val/test splits if they exist, otherwise we use $60\%$ data for training, $20\%$ data for validation (searching hyper-parameters to measure the reference transfer learning performance) and $20\%$ data for testing. Models are trained with a fixed number of epochs, and the best model in the validation split is used as the final model to be tested in the test split.

To compute the reference transfer learning performance, $\{T_m\}_{m=1}^{M}$ ($M = 12$), we carefully fine-tune pre-trained models with grid-search of hyper-parameters. \citet{li_rethinking_2020} pointed out that learning rate and weight decay are the two most important hyper-parameters. Hence, we grid-search the learning rate and weight decay (seven learning rates from $10^{-1}$ to $10^{-4}$, and seven weight decays from $10^{-6}$ to $10^{-3}$, all logarithmically spaced) to select the best hyper-parameter on the validation split and compute the accuracy on the test split as the reference transfer learning performance. \emph{It is noteworthy that LogME requires neither fine-tuning nor grid search.} Here we fine-tune pre-trained models to see how well the LogME values correlate with the reference transfer performance, but practitioners can straightforwardly use LogME to evaluate pre-trained models without fine-tuning.

We compare LogME against LEEP~\citep{nguyen_leep:_2020} and NCE~\citep{tran_transferability_2019}. Results of calculating the evidence using Laplace approximation~\citep{immer_scalable_2021} are not shown but are listed in the appendix, where we document that its performance is unsatisfactory. Before this paper, LEEP and NCE were the only two methods to rank PTMs without fine-tuning, and they can only rank supervised pre-trained models in classification tasks. We use LEEP, NCE, and LogME to compute scores $\{S_m\}_{m=1}^M$ by applying $12$ pre-trained models to the datasets. The correlation values $\tau_w$ between scores and fine-tuned accuracies are presented in Figure~\ref{fig:pre_train_selection}. 

We can find that LogME has consistently better correlation than LEEP, and outperforms NCE on most datasets (7 datasets out of 9 datasets). Note that LEEP and NCE even show negative correlation values in DTD~\citep{cimpoi_describing_2014}, because they rely on the relationship between classes of the pre-trained task and the target task but DTD classes (textures) are very different from ImageNet categories (objects). In contrast, LogME still performs reasonably well for DTD.

According to the interpretation of $\tau_w$ in Section~\ref{sec:how_to_measure_ranking}, correlation value $\tau_w$ can be roughly translated into $\frac{\tau_w + 1}{2}$ probability of correct comparison (concordant pairs). The smallest $\tau_w$ of LogME in Figure~\ref{fig:pre_train_selection} is around $0.5$, so the probability of a pre-trained model $\phi_A$ transferring better than $\phi_B$ is about $75\%$ if $\phi_A$ has a larger LogME. For most tasks $\tau_w$ of LogME is $0.7$ or $0.8$, so the probability of correct selection is $85\%$ or $90\%$, sufficient for practical usage.

\subsubsection{Ranking supervised pre-trained models in a regression task}
\label{sec:supervised_regression}

We now turn to an evaluation of how well LogME can assess pre-trained models for a regression task. The two prior methods (LEEP and NCE) depend on a categorical relationship between pre-trained categories and downstream categories, therefore they do not apply to regression tasks.

\begin{figure}[htbp]
  \centering
  \includegraphics[width=.65\columnwidth]{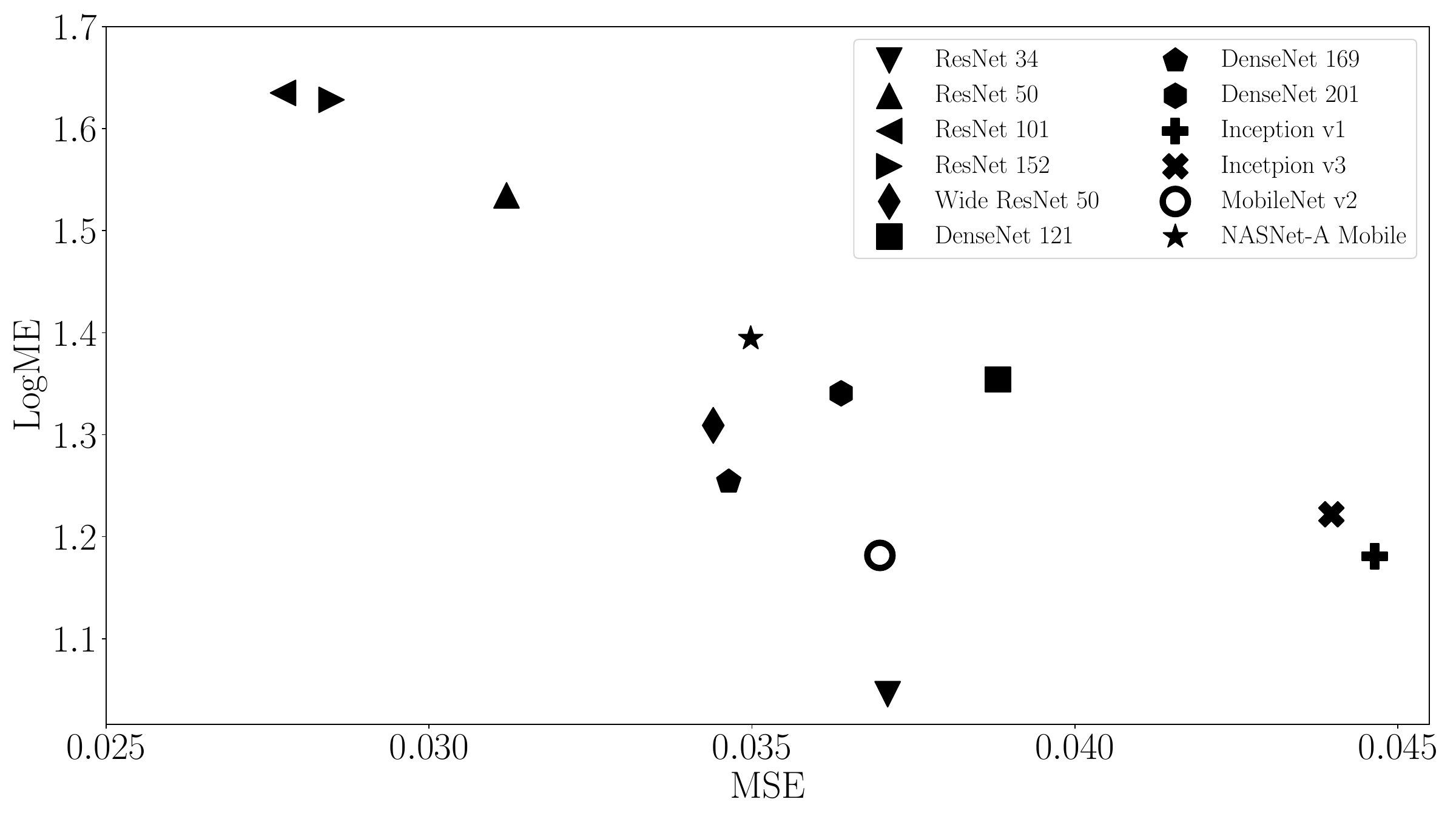}
  \vspace{-5pt}
  \caption{Supervised pre-trained models transferred to dSprites.}
  \label{fig:regression}
\end{figure}

The regression task we use is the dSprites~\citep{matthey_dsprites:_2017} dataset from the Visual Task Adaptation Benchmark~\citep{zhai_large-scale_2020}, a common benchmark for evaluating the quality of learned representations. The input is an image containing a sprite (heart, square, and ellipse) with varying scale/orientation/position. Pre-trained models are transferred to predict four scalars (scale, orientation, and $(x, y)$ positions) together, and mean square error (MSE) on the test data is reported. The supervised pre-trained models and the hyper-parameter tuning scheme are the same as in Section~\ref{sec:supervised_classification}.

Results are plotted in Figure~\ref{fig:regression}. It is clear that LogME and MSE are well correlated and the correlation coefficient $\tau_w = 0.79$ is very large: if a pre-trained model $\phi_A$ has larger LogME than $\phi_B$, with roughly $89.5\%$ probability $\phi_A$ is better (has smaller MSE) than $\phi_B$ after actually fine-tuning.

\subsubsection{Ranking contrastive pre-trained models in downstream tasks}
\label{sec:constrastive_ptms}

Unsupervised pre-trained models have attracted much attention due to their potential ability to exploit massive unlabeled datasets on the Internet~\citep{he_momentum_2020}. They use a contrastive loss~\citep{gutmann_noise-contrastive_2010} to inject supervision signals into pre-training with unlabeled data, and they feature a projection head with continuous output. Ranking contrastive pre-trained models is an important emerging challenge, and unfortunately current models such as LEEP and NCE cannot be extended to deal with the projection head of contrastive-based unsupervised pre-trained models because they rely on discrete categorical relationships.

Since LogME only requires features extracted from pre-trained models, it can be applied to contrastive pre-trained models. To demonstrate this, we use four popular models pre-trained with various training schemes: MoCo~V1~\citep{he_momentum_2020} with momentum contrast, MoCo~V2~\citep{chen_improved_2020} with an MLP projection head and strong data augmentation, MoCo~800 trained with 800 epochs as suggested by \citet{chen_simple_2020}, and SimCLR~\citep{chen_simple_2020} trained by a carefully designed training scheme~\citep{chen_simple_2020}.

For classification, we use Aircraft~\citep{maji_fine-grained_2013}, the first dataset (alphabetically) in Section~\ref{sec:supervised_classification}; for regression, we use dSprites~\citep{matthey_dsprites:_2017}, the only regression task in this paper. Results are shown in Table~\ref{tab:unsupervised}. SimCLR on dSprites is not reported as it does not converge after several trials, possibly because it is heavily tailored to classification tasks. LogME gives a \emph{perfect ordering} of both accuracy and MSE. Note that the reference order on transfer learning performance in Aircraft (MoCo~V1 $<$ MoCo~V2 $<$ MoCo 800) is different from the order in dSprites (MoCo~V1 $<$ MoCo 800 $<$ MoCo~V2), emphasizing that ranking pre-trained models is \emph{task adaptive}. We also observe that LogME values of unsupervised pre-trained models are similar (the difference is smaller than their supervised counterparts in Section~\ref{sec:supervised_classification}), mainly because unsupervised features are not very discriminative.

\begin{table}[htbp]
  \vspace{-10pt}
  \centering
  \caption{Use LogME to rank unsupervised pre-trained models.}
  \begin{small}
    \vskip 0.05in
    \begin{tabular}{ccccc}
      \toprule
    \multirow{2}[0]{*}{PTM} & \multicolumn{2}{c}{Aircraft} & \multicolumn{2}{c}{dSprites} \\
    \cmidrule(lr){2-3} \cmidrule(lr){4-5}
          &   Accuracy (\%)   &   LogME    &  MSE     & LogME \\
          \midrule
          MoCo V1 &   81.68    &   0.934    &   0.069    &  1.52 \\
          MoCo V2 &    84.16   &    0.941   &    0.047   &  1.64 \\
          MoCo 800 &    86.99   &    0.946   &   0.050    &  1.58 \\
          SimCLR &    88.10   &    0.950   &   -    &  - \\
          \midrule
          &     \multicolumn{2}{c}{$\tau_w$: 1.0}     &     \multicolumn{2}{c}{$\tau_w$: 1.0}  \\
          \bottomrule
    \end{tabular}
  \end{small}
  \label{tab:unsupervised}
\end{table}%

\subsubsection{Ranking pre-trained language models in the GLUE benchmark}
\label{sec:language_ptm_to_glue}

To further demonstrate the generality of LogME, we show how LogME can work for pre-trained language models. Again, existing methods (LEEP and NCE) cannot deal with these pre-trained language models.

\begin{figure}[htbp]
  \centering
  \includegraphics[width=\textwidth]{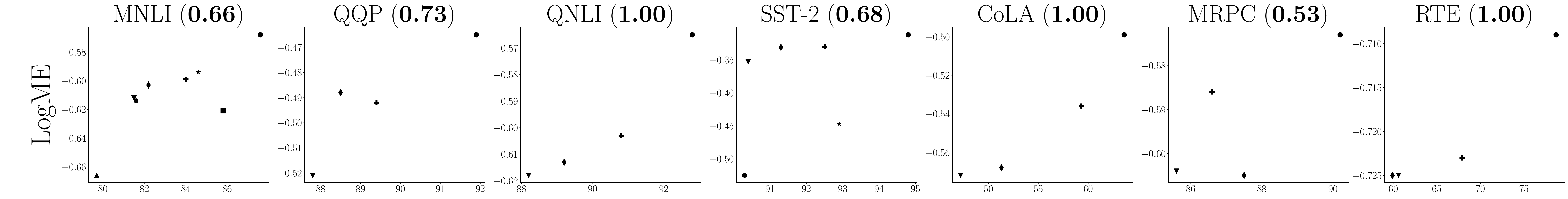}
  \textcolor{white}{\rule{\textwidth}{3pt}}
  \includegraphics[width=\textwidth]{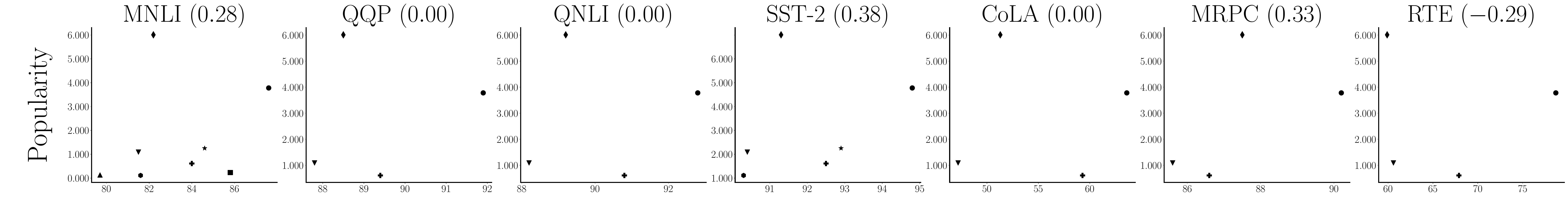}
  \includegraphics[width=.98\textwidth]{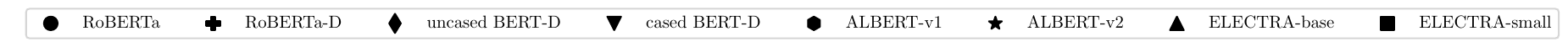}
  \vspace{-10pt}
  \caption{Correlation values ($\tau_w$) between fine-tuned accuracy (X-axis) and LogME value / Popularity value (Y-axis) in seven GLUE tasks with eight popular PTMs. One sub-figure for each task (with its $\tau_w$ in the parenthesis), and one marker for each PTM.}
  \label{fig:nlp}
  \vspace{-10pt}
\end{figure}

Here we take another approach to evaluating the reference transfer performance $\{T_m\}_{m=1}^M$. We do not fine-tune pre-trained models, but directly take fine-tuned results from \href{https://huggingface.co/models}{HuggingFace Models}, and check if LogME values can correlate well with the results. Specifically, we take pre-trained models that have GLUE performance tuned by the HuggingFace organization, and select the top eight downloaded models: RoBERTa~\citep{liu_roberta:_2019}, RoBERTa-D, uncased BERT-D, cased BERT-D, ALBERT-v1~\citep{lan_albert:_2020}, ALBERT-v2~\citep{lan_albert:_2020}, ELECTRA-base~\citep{clark_electra:_2020}, and ELECTRA-small~\citep{clark_electra:_2020} (``D'' means distilled version). The LogME values on seven GLUE tasks together with fine-tuned accuracies are plotted in Figure~\ref{fig:nlp}. Some models only have results for certain tasks and we keep them as they are. Even though these accuracy numbers are tuned by the HuggingFace organization, LogME perfectly estimates the ranking of transfer performance for three tasks (with $\tau_w = 1$), showing the surprising effectiveness of LogME in ranking pre-trained models.

One may wonder how well a pre-trained model's popularity indicates its transfer learning performance, because it is a common belief that PTMs with consistent improvements across many tasks may tend to become popular. To address this question, a quantitative measurement of popularity is required. We consider two possible quantities: the citation number of the paper proposing the pre-trained model, and the download count of the pre-trained model. The paper citation number is not a proper metric for assessing individual PTM's transferability, because one paper can contain many PTMs. For example, the BERT paper~\citep{devlin_bert:_2019} contains BERT-base and BERT-large, which have the same citation number but are in different transferability levels. Download count is a PTM-wise well-defined metric, hence we can use it as a proxy for popularity.

Thanks to the public data from \href{https://huggingface.co/models}{HuggingFace}, each PTM's download count (measured in millions) is available to approximate the popularity. The bottom figure in Figure~\ref{fig:nlp} shows how well popularity performs when it is used as a transferability metric. It is clear that popularity does not correlate well with transfer learning performance: \emph{the $\tau_w$ values of popularity are significantly lower than LogME's $\tau_w$ values}, and negative correlation values occur in the RTE task. Note that BERT models are the most popular but RoBERTa is the best among these tasks, revealing a mismatch between popularity and transfer learning performance. These experiments provide further justification for the motivation of this paper---practitioners usually select the most popular pre-trained model due to the lack of a satisfying selection strategy, and LogME can come to their rescue.

\subsubsection{Ranking pre-trained language models in a sequential tagging task}
\label{sec:ner_experiments}

So far, we have only considered simple classification and regression tasks. It would be valuable to extend LogME to tasks with structured output such as object detection and semantic segmentation. Next we show how LogME can be used in a sequential tagging task where both the input and the output are structured. How to deal with a general task with structured output is left as future work.

The specific task we consider in this section is named entity recognition~\citep{sang_introduction_2003}. It requires the model to predict the entity label (person, location, organization, etc.) of every token in a sentence, therefore the output is structured. Considering that the named entity recognition task is sometimes referred to as ``token-level classification,'' we can flatten the token dimension to apply LogME. The only change is that $n$ represents the number of tokens rather than the number of sentences.

We use the same PTMs as in Section~\ref{sec:language_ptm_to_glue}, and the dataset is CoNLL-2003~\citep{sang_introduction_2003} whose performance is measured by F-1 score. Table~\ref{tab:ner_results} holds the results. The rank correlation value $\tau_w$ is $0.20$, smaller than results in previous sections. The small $\tau_w$ is caused by an outlier PTM named RoBERTa~\citep{liu_roberta:_2019}, which has the largest F-1 score with a relatively small LogME value. We conjecture that RoBERTa has a small LogME value because it is trained much longer than BERT in the masked language modeling task, which might make its representation tailored to the task, lowering its LogME score in the dissimilar task of named entity recognition. On the other hand, RoBERTa is robustly optimized, so it can be easily fine-tuned to downstream tasks with competitive results.

If we select the best PTM by the largest LogME value, ALBERT-v1 will be used and its performance is comparable to the best ($97.0\%$ \emph{vs.} $97.4\%$). From this perspective, LogME is reasonably useful. In general, how to deal with structured tasks better is a research problem requiring further effort.

\begin{table}[h]
\addtolength{\tabcolsep}{-4pt}
\centering
\caption{Ranking pre-trained models in named entity recognition (CoNLL-2003 task).}
\vskip 0.05in
\resizebox{\textwidth}{!}{
  \begin{tabular}{cccccccccc}
    \toprule
  PTM & RoBERTa & RoBERTa-D & uncased BERT-D & cased BERT-D & ALBERT-v1 & ALBERT-v2 & ELECTRA-base & ELECTRA-small & $\tau_w$ \\
        \midrule
  F-1 score ($\%$) & 97.4 & 96.6 & 96.8 & 95.5 & 97.0 & 97.4 & 97.2 & 91.9 &  \\
  \midrule
  LogME & 0.685 & 0.723  & 0.783  &   0.623   &  0.834  &   0.809  &  0.746   &  0.646 &  0.20 \\
  \bottomrule
  \end{tabular}%
}
\label{tab:ner_results}%
\end{table}%

\subsection{Tuning pre-trained models}
\label{sec:exp_tuning_ptms}

This section turns to the second part of the proposed paradigm: tuning pre-trained models. As mentioned in Section~\ref{sec:tuning_ptms}, most academic researchers are not constrained by the inference cost of deployed models, and they can use the best-ranked (according to the LogME value) PTM straightforwardly. This paper is concerned with the practical usage scenario, where computational constraints require us to use a specific PTM but we still want to leverage the knowledge from other PTMs in the pre-trained model hub.

The experiments in this section are designed to compare three methods of tuning multiple PTMs: the knowledge distillation approach, the Zoo-tuning approach and the proposed B-Tuning method. We first conduct experiments with multiple homogeneous PTMs where all three methods are applicable, then we dive into the practical case of multiple heterogeneous PTMs. By default, the temperature scaling hyper-parameter $t$ is set to $0.1$ in Equation~\ref{LBayesian}.

\subsubsection{Tuning multiple homogeneous PTMs}
\label{sec:tuning_homo_ptms}

We use five homogeneous pre-trained models following the experimental setup of Zoo-tuning~\citep{shu_zoo-tuning:_2021}.  They are ResNet-50 models trained by different pre-training tasks: (1) Supervised pre-trained on ImageNet~\citep{he_deep_2016}; (2) Unsupervised pre-trained by MoCo~\citep{he_momentum_2020}; (3) MaskRCNN model~\citep{he_mask_2017}; (4) DeepLab V3~\citep{chen_rethinking_2017}; (5) KeyPoint detection model pre-trained on COCO~\citep{lin_microsoft_2014}. The dataset we use is Aircraft~\citep{maji_fine-grained_2013}, the first dataset (alphabetically) in Section~\ref{sec:supervised_classification}. The target model is ResNet-50 pre-trained in ImageNet, following the setting of \citet{shu_zoo-tuning:_2021}.

\emph{To demonstrate the effectiveness of B-Tuning,} we use all five PTMs as the teacher models, and report the performance of three methods (B-Tuning, knowledge distillation, and Zoo-tuning) on multiple PTM tuning in the first row of Table~\ref{tab:multiple_ptm_tuning}. Zoo-tuning performs better than vanilla knowledge distillation, but the new B-Tuning method surpasses Zoo-tuning, setting a new state-of-the-art benchmark for multiple PTM tuning.

\begin{table}[htbp]
  \centering
  \vspace{-10pt}
  \caption{Accuracy ($\%$) of multiple PTM tuning in Aircraft, with different teacher models and tuning methods. As a baseline, single PTM fine-tuning yields $82.99\%$ accuracy.}
  \begin{small}
    \begin{tabular}{|l|c|c|c|}
      \hline
        \diagbox{teacher models}{method}  & Knowledge Distillation & Zoo-tuning & B-Tuning \\
        \hline
    all PTMs from the PTM hub &   $82.97\pm 0.27$    &   $83.32\pm 0.32$    & $83.49\pm 0.17$ \\
    \hline
    top-3 PTMs (ranked by LogME) &    $84.29\pm 0.30$   &    -   & $85.12\pm 0.15$ \\
    \hline
    \end{tabular}%
  \end{small}
  \label{tab:multiple_ptm_tuning}%
\end{table}%

\emph{To demonstrate the effectiveness of LogME selection in multiple PTM tuning,} we rank five PTMs by LogME, and select the top-$K$ PTMs as the teacher models in the subsequent tuning. \citet{shu_zoo-tuning:_2021} used all five PTMs to tune the target PTM, since they do not investigate how to select PTMs. To sufficiently test the effect of selection, we choose $K = \arg \max_{3 \le K \le 5} (_K^5) = 3$, so that there are many possible selections and later we can explore how optimal LogME selection is. The results are in the second row of Table~\ref{tab:multiple_ptm_tuning}. Surprisingly, selecting top-3 PTMs brings a significant performance improvement, demonstrating the effectiveness of the \emph{``ranking and tuning pre-trained models''} paradigm.

\begin{figure}[htbp]
  \hspace{.265\columnwidth}\includegraphics[width=.735\columnwidth]{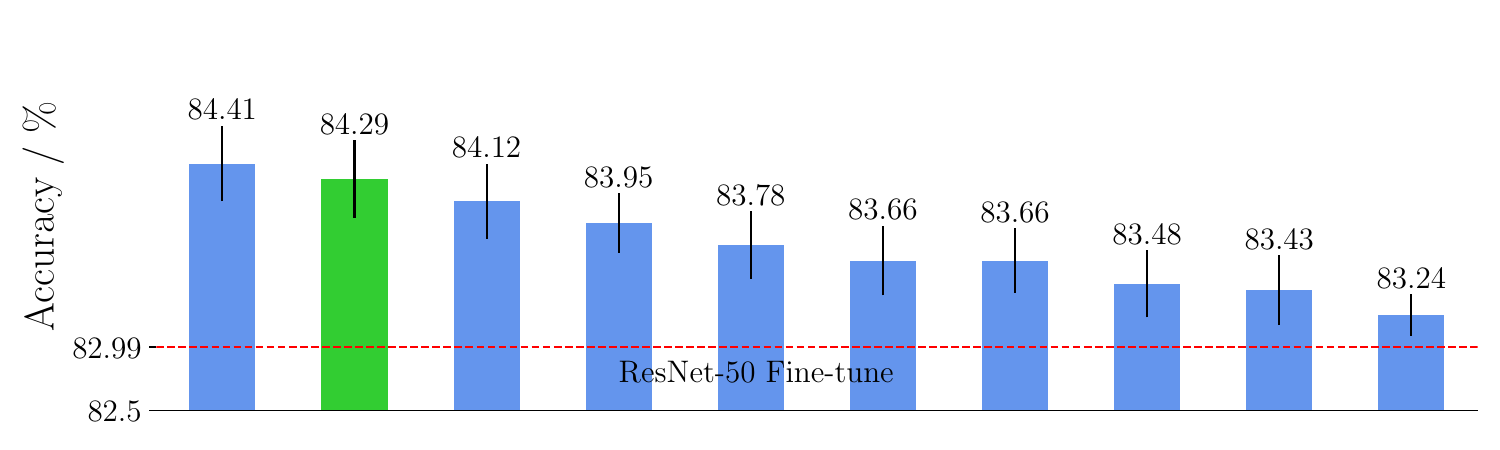}
  \centering
  \begin{small}
  \begin{tabular}{p{80pt}c p{10pt} p{16pt}p{16pt}p{16pt}p{16pt}p{16pt}p{16pt}p{16pt}p{16pt}p{16pt}p{16pt}}
      \toprule
      {ImageNet Sup.} & {0.947} & \vline & \checkmark & {\checkmark} & \checkmark &            & \checkmark & \checkmark &            &            &            & \checkmark \\
      {MaskRCNN PT.} & {0.936}  & \vline & \checkmark & {\checkmark} &            & \checkmark &            & \checkmark &            & \checkmark & \checkmark &            \\ 
      {MoCo PT.} &  {0.934}     & \vline &            & {\checkmark} & \checkmark & \checkmark & \checkmark &            & \checkmark & \checkmark &            &            \\ 
      KeyPoint PT. & 0.914                                    & \vline &            &                             &            &            & \checkmark & \checkmark & \checkmark & \checkmark & \checkmark & \checkmark \\
      DeepLab PT. & 0.913                                     & \vline & \checkmark &                             & \checkmark & \checkmark &            &            & \checkmark &            & \checkmark & \checkmark \\  
      \midrule
      PTM name & {\scriptsize LogME} & \vline & \multicolumn{10}{c}{} \\ \bottomrule
  \end{tabular}
  \end{small}
  \caption{Accuracy of knowledge distillation with three PTM teachers. All $(_3^5)=10$ combinations of selecting $3$ PTM teachers are reported. Selecting top-3 PTMs according to LogME achieves the second best performance among ten combinations.}
  \vspace{-10pt}
  \label{fig:homo}
\end{figure}

\emph{To evaluate the optimality of LogME selection,} we try all the $(_3^5)=10$ combinations of selecting three PTMs from five PTMs. Vanilla knowledge distillation is used to avoid confounders. Results are shown in Figure~\ref{fig:homo}, with the accuracy of fine-tuning a single ResNet-50 as the baseline. We have two observations from Figure~\ref{fig:homo}: (1) Transferring the knowledge from multiple PTMs consistently outperforms fine-tuning a single pre-trained model ($82.99\%$), which adheres to our intuition that utilizing the rich knowledge from various PTMs is better than fine-tuning alone. (2) The best combination achieves $84.41\%$ accuracy, but usually it is too expensive to try all the combinations (10 trials). Instead, we can use LogME to select the top-3 PTMs, which achieves $84.29\%$ accuracy and is the second best. Moreover, we can select the top-3 PTMs by LogME and then perform B-Tuning, which even surpasses the best combination and has an accuracy of $85.12\%$.

We can draw three conclusions from experiments in this section: (1) multiple PTM tuning is better than single PTM fine-tuning; (2) it is better (near-optimal among all the possible selections) to select top-ranked PTMs according to LogME than to use all the PTMs; (3) B-Tuning is superior to knowledge distillation and Zoo-tuning.

It is important to point out that selection based on LogME value is a greedy procedure, and this procedure could fail to capture complicated high-order interactions among PTMs. For example, in Figure~\ref{fig:homo}, DeepLab pre-trained model has the lowest LogME value, but it appears in the best combination. How to analyze the high-order interactions among PTMs would be a worthwhile research question in the future.

\subsubsection{Tuning multiple heterogeneous PTMs}
\label{sec:tuning_hetero_ptms}

Section~\ref{sec:tuning_homo_ptms} studies multiple PTM tuning with homogeneous models, which follows the setting of \citet{shu_zoo-tuning:_2021} and demonstrates the superiority of B-Tuning. Nonetheless, compared with tuning multiple homogeneous PTMs, a more general and more attractive application of multiple PTM tuning is to transfer knowledge from a large hub of heterogeneous PTMs. This section focuses on the latter setting, and provides some guidelines on how to select a proper number of PTMs (\emph{i.e.}, the hyper-parameter $K$) as teachers. 

\begin{figure}[htbp]
  \centering
  \includegraphics[width=.8\columnwidth]{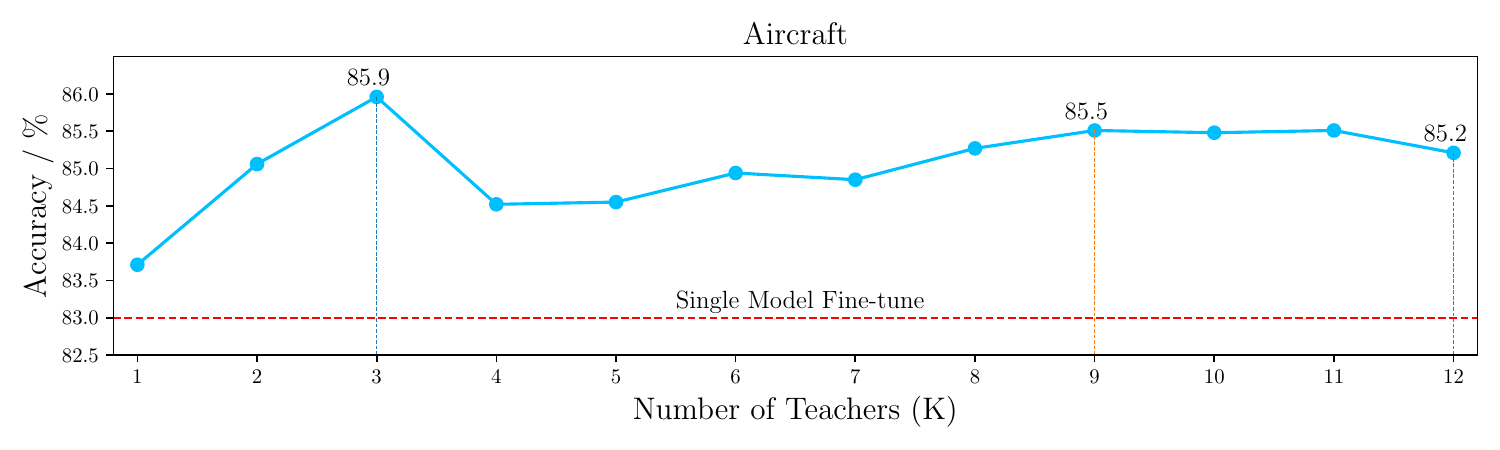}
  \includegraphics[width=.8\columnwidth]{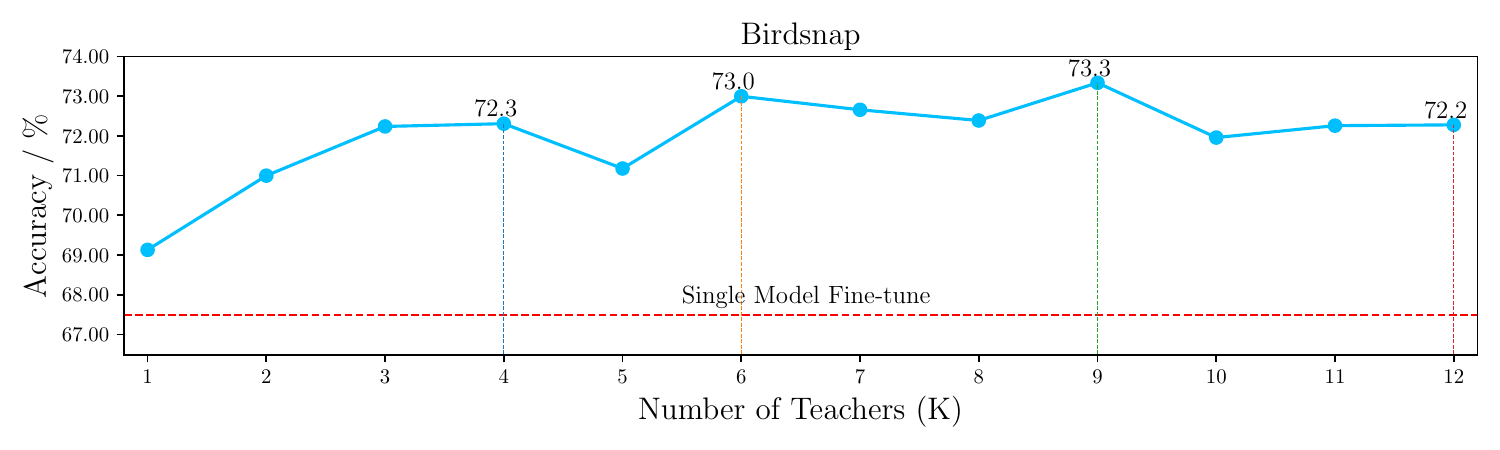}
  \vspace{-10pt}
  \caption{A study on the number of PTMs ($K$) to use in B-Tuning with two datasets (Top: Aircraft; Bottom: Birdsnap).}
  \label{fig:hetero}
\end{figure}

The alphabetically first and second datasets (Aircraft and Birdsnap) are chosen and the PTM hub consists of the $12$ PTMs used in Section~\ref{sec:supervised_classification}. The $12$ PTMs are ranked by their LogME values, and the target PTM $\phi_t$ is the most common ResNet-50. Top-$K$ PTMs are used in B-Tuning to fine-tune the target model, with $K$ varying from $1$ to $12$. Results are plotted in Figure~\ref{fig:hetero}, where the X-axis is the value of $K$.

We make the following observations based on Figure~\ref{fig:hetero}: (1) B-Tuning with multiple PTMs is consistently better than single PTM fine-tuning. (2) B-Tuning with all the $12$ PTMs does not yield the best accuracy, which emphasizes the importance of selecting proper PTMs. (3) The trend of accuracy with respect to $K$ is complicated, and how to select an optimal $K$ is a worthwhile topic for future research.

For practitioners, there are two concerns about the choice of $K$: (1) choosing the optimal $K$ can yield the best accuracy; (2) but larger $K$ incurs a much larger computational cost, since a forward pass of each PTM during tuning is required. Considering the results in Figure~\ref{fig:hetero} and the trade-off between the computational cost and the performance improvement, we recommend choosing $K$ from $\{2, 3, 4\}$ in practice.

\subsection{Using ImageNet-1K as the downstream task}

The above experiments focus on small-scale and medium-scale downstream tasks, which are common in transfer learning research. This section takes a step further to use the large-scale ImageNet-1K~\citep{deng_imagenet:_2009} as the downstream dataset. In this case, a dataset larger than ImageNet-1K should be used for pre-training. JFT-300M~\citep{sun_revisiting_2017}, Instagram-1B~\citep{mahajan_exploring_2018}, and ImageNet-21K~\citep{deng_imagenet:_2009} are commonly-used datasets that are larger than ImageNet-1K. Among them, ImageNet-21K is the only publicly available dataset, which serves as the pre-training dataset here. ImageNet-21K pre-trained models are provided by the \href{https://github.com/rwightman/pytorch-image-models}{timm} project. It mainly contains models pre-trained on ImageNet-1K, but also has three models pre-trained on ImageNet-21K and fine-tuned on ImageNet-1K, including MLP-Mixer~\citep{tolstikhin_mlp-mixer:_2021}, ViT~\citep{dosovitskiy_image_2021}, and Swin-T~\citep{liu_swin_2021}. With ImageNet-1K as the downstream dataset, their LogME score and fine-tuned reference transfer learning performance are presented in Table~\ref{tab:imagenet_results}. LogME is perfectly aligned with the reference performance, with a correlation value $\tau_w = 1$. Then we use B-Tuning to tune the commonly used ResNet-50 trained in ImageNet-1K, with ViT and Swin-T as teacher models. The accuracy is increased from $76.15\%$ to $76.50\%$.

Experiments in this section demonstrate that LogME and B-Tuning work for not only small-scale and medium-scale datasets but also for large-scale datasets.

\begin{table}[h]
\addtolength{\tabcolsep}{-3pt}
\centering
\caption{Ranking models pre-trained on ImageNet-21K transferred to ImageNet-1K.}
\vskip 0.05in
  \begin{tabular}{cccccccccc}
    \toprule
  PTM pre-trained on ImageNet-21K & MLP-Mixer & ViT & Swin-T & $\tau_w$ \\
        \midrule
  Fine-tuned Accuracy on ImageNet-1K ($\%$) & 76.61 & 84.53 &  85.25 & \multirow{2}[0]{*}{1.00} \\
  \cline{1-4}
  LogME value & 2.075 & 2.085 & 2.134  &   \\
  \bottomrule
  \end{tabular}%
\label{tab:imagenet_results}%
\end{table}%

\subsection{Efficiency of LogME}
\label{sec:timecost}

A theoretically sound algorithm is often complex and computationally expensive, and this is the case for LogME without optimization. Fortunately, we successfully reduced the computational complexity after analyzing the theoretical convergence of LogME by the fixed point iteration (see Section~\ref{sec:convergence_analysis}). We present a summary of the algorithmic complexity in Table~\ref{tab:optimization}, and present empirical results in Table~\ref{tab:speedup}, where we show the wall-clock-time speedup measured in Aircraft with ResNet-50. The na\"ive implementation is very slow. Our conference paper~\citep{you_logme:_2021} proposed an optimization scheme for matrix multiplication and matrix inversion, which brings $61.7\times$ speedup. This paper further proposes the fixed point iteration algorithm, which results in a much larger speedup ($131.5\times$). Thanks to the optimized method, LogME is not only theoretically sound but also computationally efficient.

\begin{table}[htbp]
  \centering
  \caption{Quantitative measurement of computational speedup in evidence maximization.}
  \vskip 0.05in
  \resizebox{\columnwidth}{!}{
    \begin{tabular}{lcc}
      \toprule
          & Wall-clock time (second) & Speedup \\
          \midrule
    evidence maximization (na\"ive implementation) &    $802.5 \pm 5.6$   & - \\
    evidence maximization (optimized by \citet{you_logme:_2021}) &    $13.1 \pm 0.7$   & $61.7 \times$ \\
    evidence maximization (fixed point iteration, proposed) &    $6.1 \pm 0.7$   & $131.5 \times$  \\
    \bottomrule
    \end{tabular}}
  \label{tab:speedup}
\end{table}%

Next, we quantitatively measure the wall-clock and memory footprint of LogME in both computer vision and natural language processing; see Table~\ref{tab:efficiency}. ResNet~50 on Aircraft is used for computer vision, and RoBERTa-D on MNLI task is used for NLP. The cost for the rest of the models and datasets varies, but the proportion is similar. The cost of computing reference transferability $T_m$ (fine-tuning with hyper-parameter search) serves as the upper bound of ranking pre-trained models. Note that, because carelessly tuned hyper-parameters cannot tell good models apart from bad models, it is necessary to attribute the cost of hyper-parameter search to fine-tuning. We also list the cost of extracting features by pre-trained models, which is the lower bound of ranking pre-trained models.

\begin{table}[htbp]
  \centering
  \caption{Computational cost and memory footprint of LogME.}
  \vskip 0.05in
  \resizebox{\columnwidth}{!}{
    \begin{tabular}{c|l|l}
          \toprule
          & wall-clock time & memory footprint \\
          \hline
    \multirow{4}[0]{*}{Computer Vision} &  fine-tune (upper bound) \hfill $161000$s    & fine-tune (upper bound) \hfill 6.3 GB\\
    &    extract feature (lower bound) \hfill $37$s  & extract feature (lower bound) \hfill 43 MB \\
          &    LogME \hfill $43$s  & LogME \hfill 53 MB \\
          &     benefit \hfill $3700\uparrow$ & benefit  \hfill $120\uparrow$  \\
          \hline
    \multirow{4}[0]{*}{Natural Language Processing} &      fine-tune (upper bound)  \hfill 100200s   & fine-tune (upper bound)\hfill 88 GB \\
    &    extract feature (lower bound) \hfill $1130$s  & extract feature (lower bound) \hfill 1.2 GB \\

          &    LogME \hfill 1136s  & LogME \hfill 1.2 GB \\
          &    benefit \hfill $88\uparrow$  &  benefit \hfill $ 73 \uparrow$\\
          \bottomrule
    \end{tabular}}
  \label{tab:efficiency}
\end{table}%

Based on Table~\ref{tab:efficiency}, we have the following observations: (1) brute-force fine-tuning is computationally expensive, requiring about a day for one dataset with one pre-trained model. Selecting the best pre-trained model out of $12$ models would cost $12$ GPU-days. (2) Extracting features is very cheap and costs much less than fine-tuning. (3) The additional time-cost of LogME compared to feature extraction is rather small, which means that \emph{LogME's cost is very close to the lower bound}. In computer vision, LogME is $3700\times$ faster than fine-tuning, with $120\times$ less memory footprint. In the NLP domain, feature extraction is much slower than that in computer vision, and therefore the wall-clock time speedup ($88\times$) is not that striking.

In summary, LogME is efficient in terms of both wall-clock time and memory footprint, thanks to the optimized algorithm (fixed point iteration) inspired by the theoretical analysis.

\subsection{Comparing LogME to re-training head}
\label{sec:compare_head}

A straightforward way to measure the relationship between features and labels is to train a linear classification/regression head for the downstream task, and to use the head's performance as a metric, which is known as ``\emph{linear probing}'' or ``\emph{linear protocol evaluation}.'' Empirically, we find that re-training the head does not work well. In the following, we summarize why re-training the head is inferior to LogME from three perspectives, which partially explains why the important problem of ranking and tuning PTMs was under-explored in the past. 

\begin{figure}[htbp]
  \centering
  \vspace{-10pt}
  \includegraphics[width=.7\columnwidth]{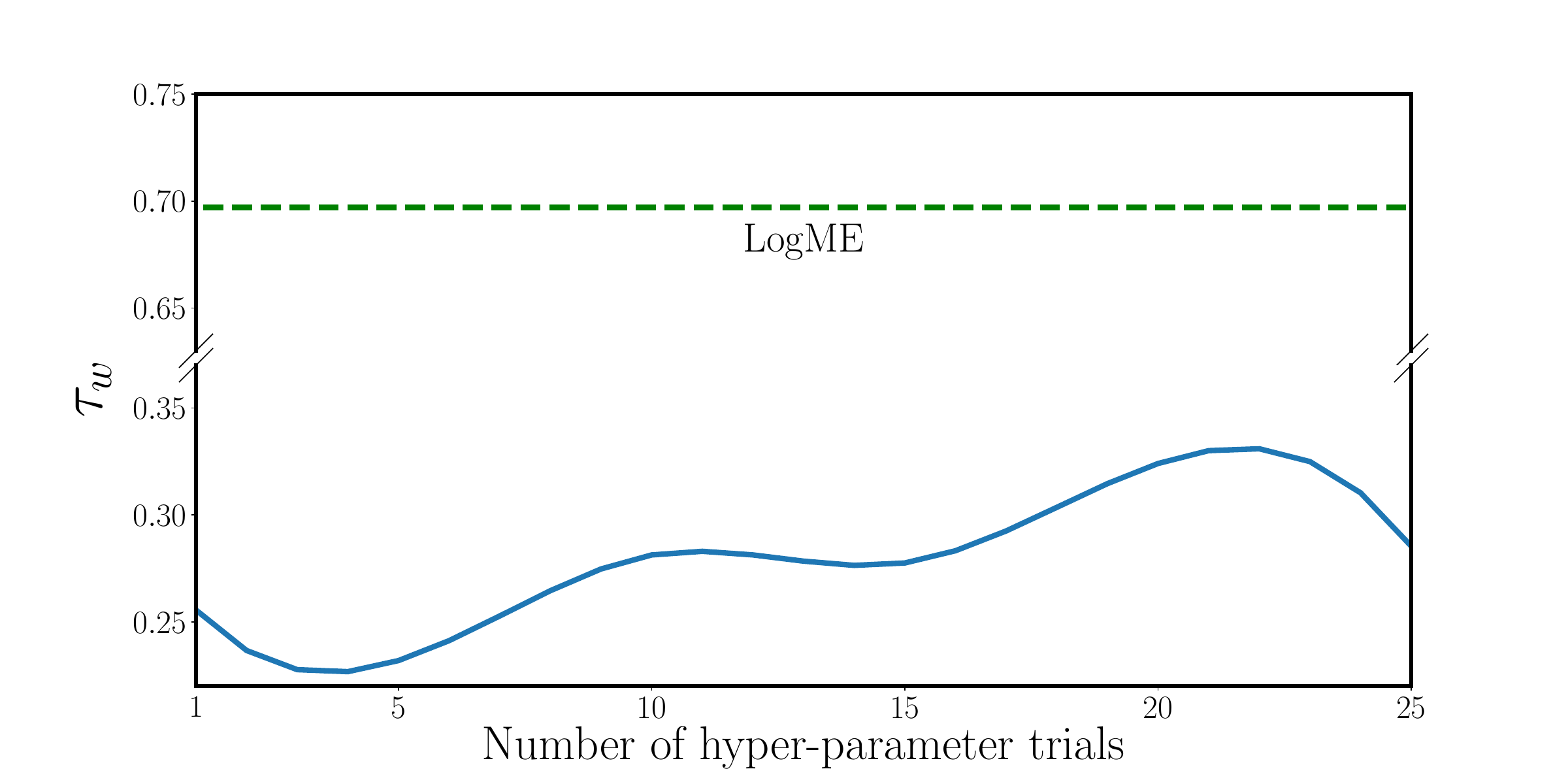}
  \vspace{-10pt}
  \caption{The correlation of re-training the head with respect to the number of hyper-parameter trials.}
  \vspace{-10pt}
  \label{fig:search_space}
\end{figure}

\textbf{(1) LogME is more efficient than re-training head.} In linear protocol evaluation, parameters in the head are learned by maximum likelihood estimation, which is prone to over-fitting. To alleviate over-fitting, grid search for its hyper-parameters (such as the strength of L2 regularization) should be tuned extensively on a validation set, making head re-training inefficient. For example, in the Caltech dataset, we extract features from 12 PTMs, train softmax regressors with tuned hyper-parameters (the L2 regularization strength), and plot the correlation $\tau_w$ between the best head accuracy and the reference transfer performance with respect to the number of hyper-parameter trials in Figure~\ref{fig:search_space}. The correlation of LogME is plotted as a reference. Computing LogME requires $3\times$ less time than re-training a head with one fixed hyper-parameter, and re-training a head with exhaustive hyper-parameter search is still much inferior to LogME.

\textbf{(2) Re-training head does not work well with limited training data.} Because re-training the head follows a supervised learning paradigm, it suffers in low-shot learning scenarios. For example, prompt learning~\citep{liu_pre-train_2021} is an active research area in natural language processing, where researchers try to exploit the potential of frozen pre-trained models with only a few training data points. In the sentiment classification task SST-2~\citep{socher_recursive_2013}, prompt learning~\citep{liu_pre-train_2021} extracts the sentence embedding $E_S$ for each sentence $S$, and compares $E_S$ with word embeddings $E_P, E_N$ of a positive anchor word $P$ (such as ``good'' and ``fantastic'') and a negative anchor word $N$ (such as ``bad'' and ``awful''). The decision rule is: \texttt{sentence S contains positive sentiment $\iff E_S^T E_P > E_S^T E_N$}. In this case, searching for proper anchor words on validation data yields $61.4\%$ accuracy (complete results are available in Table~\ref{tab:prompt_learning}). Re-training the head (\emph{i.e.}, training a simple classification head with limited training data on the frozen sentence embedding $E_S$ and tuning the weight-decay hyper-parameter on a validation set) only achieves $51.64\%$ accuracy with $10$ sentences for training. Meanwhile, we can apply LogME (or to be specific, the posterior predictive distribution introduced in Section~\ref{sec:bayesian_tuning}) to this problem, which does not require any hyper-parameter tuning. This way, we can combine training data with validation data to compute the predictive weight $m$ for each class, and use $m$ as the embedding of a virtual ``anchor word,'' which results in $79.24\%$ accuracy, \emph{a huge improvement over re-training head and manually selected anchor words}.  Moreover, the superior performance of LogME is interpretable: we analyzed the predictive weight $m$ for the negative sentiment class, and find that it is closest to the embedding of ``\texttt{dump},'' ``\texttt{:(},'' ``\texttt{doomed},'' ``\texttt{Worse},'' and ``\texttt{worse}.'' Interestingly, it can discover that ``\texttt{:(},'' a cyber word used to express unhappy emotion, contains negative sentiment.

\textbf{(3) Re-training the head does not have a clear metric.} As a side issue, even if we re-train a head for a downstream task, it is unclear which quantity should be used as the ranking metric. When the performance of a downstream task is evaluated by accuracy or MSE, is it over-fitting to use the accuracy or MSE of the re-trained head? Indeed, in Figure~\ref{fig:search_space}, when the number of hyper-parameter trials increases, the correlation can even go down, confirming the concern of over-fitting. In contrast, LogME is based on unified modeling of label evidence, which has clear statistical support.

\section{Conclusions}
\label{sec:conclusion}

Pre-trained models are universally acknowledged as a foundation of deep learning. Researchers have explored many ways to create and exploit PTMs. In this paper, we change the focus from individual PTMs to PTM hubs, and study how to sufficiently exploit PTM hubs within a new paradigm of ranking and tuning pre-trained models. The ranking part introduces a theoretically sound and computationally efficient transferability metric named LogME. LogME is then further extended to be a multiple PTM tuning method that we refer to as B-Tuning, which completes the tuning part of the paradigm.

We presented extensive experiments that confirm the effectiveness of the proposed methods in ranking (LogME \emph{vs.}\ brute-force fine-tuning/LEEP/NCE), selection (top-$K$ PTMs by LogME \emph{vs.}\ exponentially many combinations), and tuning (B-Tuning \emph{vs.}\ Zoo-tuning and Knowledge Distillation), showing that the new paradigm of exploiting PTM hubs is attractive for practitioners. 

\acks{We would like to present our thanks to Ximei Wang, Xinyang Chen, Yang Shu at Tsinghua University, Yi Zeng at Peking University, and Yonglong Tian at MIT for helpful discussions.
Kaichao You, Yong Liu, Jianmin Wang, and Mingsheng Long are supported by the National Key Research and Development Project (2021YFB1715200), the National Natural Science Foundation of China (62022050 and 62021002), the Beijing Nova Program (Z201100006820041), the BNRist Scholar Fund (BNR2021RC01002), and the Tsinghua-Huawei Innovation Fund.}

\bibliography{example_paper}

\newpage 
\appendix

\section{Notation Table}

Notation used in this paper is listed in the following table. We endeavored to avoid notational conflicts, but the following  conflicts are worth noting: (1) $f_i$ represents features of $x_i$ and $f(t)$ represents the fixed point iteration function. (2) $w$ represents parameters in the linear head while the weighted Kendall's rank correlation $\tau_w$ uses $w$ as its subscript. (3) $t$ is used in the convergence proof and also used for the temperature hyper-parameter in B-Tuning.

\begin{table}[htbp]
  \centering
  \caption{Notations used in this paper.}
  \vskip 0.05in
  \resizebox{!}{.35\textheight}{
    \begin{tabular}{ccl}
      \toprule
      notation & dimensionality & meaning \\
      \hline
      $i, j, k$     & $\mathbb{N}$ & running subscripts     \\ \hline
      $M , n$    & $\mathbb{N}$  & the number of PTMs and samples \\ \hline
      $K$    & $\mathbb{N}$  & the number of selected PTMs for subsequent tuning \\ \hline
      $C , D$   & $\mathbb{N}$   & the dimension of label and extracted feature    \\ \hline
      $\phi$   & $-$   & a pre-trained model  \\ \hline 
      $x_i$   & $-$   & an input sample     \\ \hline 
      $f_i = \phi (x_i)$   & {\scriptsize $\mathbb{R}^D$}   & extracted feature of an input example \\ \hline
      {\scriptsize $F=[f_1, \ldots, f_n]^T$}   & {\scriptsize $\mathbb{R}^{n\times D}$}   & stacked features of $f_i$ \\ \hline
      $Y_i$   & {\scriptsize $\mathbb{R}^C$}   & label of $x_i$ \\ \hline
    $y_i$   & $\mathbb{R}$   & a component of $Y_i$ \\ \hline
    $y$    & $\mathbb{R}^n$  & the label component for all $n$ samples     \\ \hline
    $y'$    & $-$  & predictive distribution of an input sample \\ \hline
    $\pi_k = \frac{\exp (\mathcal{L}_{k} / t)}{\sum_{j=1}^K \exp (\mathcal{L}_{j} / t)}$    & $\mathbb{R}$  & weighted coefficient for each teacher model $\phi_k$ \\ \hline
    {\scriptsize $\bar{y}' = \sum_{k=1}^K \pi_k y_{k}'$}   & $-$  & weighted average of $y'$     \\ \hline 
    $T$    & $\mathbb{R}$  & reference transfer performance \\ \hline
    $S$    & $\mathbb{R}$  & score produced by a transferability metric     \\
    \hline 
    $\tau$   & $\mathbb{R}$   & Kendall's rank correlation \\ \hline
    $\tau_w $   & $\mathbb{R}$   & weighted Kendall's rank correlation \\ \hline
    $w$    & {\scriptsize $\mathbb{R}^D$ } & parameter in linear head     \\
    \hline 
    $\alpha, \beta$   & $\mathbb{R}$   & hyper-parameter of the Bayesian linear model \\ \hline
    {\scriptsize $A = \alpha I_D + \beta F^TF$}   & {\scriptsize $\mathbb{R}^{D\times D}$}   & a quantity in calculating LogME \\ \hline
    {\scriptsize $m = \beta A^{-1}F^Ty$}   & {\scriptsize $\mathbb{R}^D$}   & a quantity in calculating LogME     \\ \hline
    $\gamma$   & $\mathbb{R}$   & a quantity in calculating LogME \\ \hline 
    {\scriptsize $\mathcal{L} = \mathcal{L}(\alpha, \beta)$}  & $\mathbb{R}$ & log evidence given $\alpha, \beta$ \\ \hline
    $\alpha^*, \beta^*$   & $\mathbb{R}$   & $\alpha, \beta$ to achieve maximum evidence     \\
    \hline 
    $U, \Sigma, V$   & $-$   & matrices in SVD ({\scriptsize $F = U \Sigma V^T$})     \\
    \hline 
    $\sigma$   & $\mathbb{R}$   & diagonal entries in $\Sigma$ \\ \hline
    $r$     & $\mathbb{N}$ & the rank of matrix $F$    \\ \hline
    {\scriptsize $z = U^Ty$}    & $\mathbb{R}^n$  & transformation of $y$ under $U$ \\ \hline
    $t=\frac{\alpha}{\beta}$   & $\mathbb{R}$   & a quantity in convergence proof     \\
    \hline 
    $t', \alpha', \beta'$    & $\mathbb{R}$  & the value of $t, \alpha, \beta$ after an iteration \\ \hline
    $f(t)$  & $-$    & the fixed point iteration function     \\
    \hline 
    {\scriptsize $\tilde{\mathcal{L}}, \tilde{F}$}  & $-$    & ${\mathcal{L}}, {F}$ for duplicated or padded features \\ \hline
    $W$    & $-$  & transformation matrix in knowledge distillation     \\
    \bottomrule
    \end{tabular}
  }
  \label{tab:notations}%
\end{table}%
\newpage

\section{Proof of Theorem~\ref{thm:f}}
\label{proof:f}

Theorem~\ref{thm:f}: \textup{Algorithm~\ref{alg:iteration} induces a scalar function} $t' = f(t) = \left(\frac{n}{n - \sum_{i=1}^{D} \frac{ \sigma_i^2}{t + \sigma_i^2}} - 1\right) t^2 \frac{\sum_{i=1}^{n} \frac{z_i^2}{(t + \sigma_i^2)^2}}{\sum_{i=1}^{n} \frac{\sigma_i^2 z_i^2}{(t + \sigma_i^2)^2}}$. \\

\begin{proof}
  Let's express all symbols in a unified form with respect to $\alpha, \beta, \Sigma, z, U, V$:

\begin{itemize}
  \item $A = \alpha I + \beta F^TF = V (\alpha I + \beta \Sigma^T\Sigma) V^T$
  \item $A^{-1} = V \Sigma_{inv} V^T$ where $(\Sigma_{inv})_{ii} = \frac{1}{\alpha + \beta \sigma_i^2} \; (1 \le i \le D)$
  \item $m = \beta A^{-1}F^Ty = \beta V \Sigma_{inv} \Sigma^T z$
  \item $m^Tm = z^T \Sigma_{m} z$ with $\Sigma_m = \beta^2\Sigma \Sigma_{inv}^2 \Sigma^T$ and $(\Sigma_m)_{ii} = \frac{\beta^2 \sigma_i^2}{(\alpha + \beta \sigma_i^2)^2}$, so $m^Tm = \sum_{i=1}^{n} \frac{\beta^2 \sigma_i^2 z_i^2}{(\alpha + \beta \sigma_i^2)^2}$
  \item $Fm = \beta U \Sigma \Sigma_{inv} \Sigma^T z$, $Fm - y = U \Sigma_{res} z$ with $\Sigma_{res} = \beta \Sigma \Sigma_{inv} \Sigma^T - I$, $(\Sigma_{res})_{ii} = - \frac{\alpha}{\alpha + \beta \sigma_i^2}$
  \item $\vert \vert F m - y \vert \vert_2^2 = (Fm-y)^T(Fm-y) = z^T (\Sigma_{res})^2 z = \sum_{i=1}^{n} \frac{\alpha^2 z_i^2}{(\alpha + \beta \sigma_i^2)^2}$
  \item $\gamma = \sum_{i=1}^{D} \frac{\beta \sigma_i^2}{\alpha + \beta \sigma_i^2} = \sum_{i=1}^{D} \frac{ \sigma_i^2}{t + \sigma_i^2}$
\end{itemize}

Putting them together, we have
\begin{equation*}
  t' = \frac{\alpha'}{\beta'} = \frac{\gamma}{n - \gamma} \frac{\vert \vert F m - y \vert \vert_2^2}{m^Tm} = \left(\frac{n}{n - \sum_{i=1}^{D} \frac{ \sigma_i^2}{t + \sigma_i^2}} - 1\right) t^2 \frac{\sum_{i=1}^{n} \frac{z_i^2}{(t + \sigma_i^2)^2}}{\sum_{i=1}^{n} \frac{\sigma_i^2 z_i^2}{(t + \sigma_i^2)^2}} = f(t)
\end{equation*}
\end{proof}

\vspace{-20pt}
\section{Proof of Theorem~\ref{thm:converge}}
\label{proof:converge}

Theorem~\ref{thm:converge}: \textup{If} $r < n$ \textup{and} $\sum_{1 \le i,j \le n}(z_i^2-z_j^2)(\sigma_i^2-\sigma_j^2)>0$, \textup{then} $f(t)$ \textup{has a fixed point and thus MacKay's algorithm will converge.} \\

\begin{proof}
  The theorem can be proved by studying the behavior of $f(t)$ near $0$ and $\infty$.

  We have $\lim_{t\to 0}f(t)=\frac{r}{n-r}\frac{\sum_{i=r+1}^n z_i^2}{\sum_{i=1}^r z_i^2} > 0$, which is a constant and positive number.
  
  When $t$ approaches infinity, we find that $\lim_{t\to \infty}\frac{f(t)}{t}=\frac{\sum_{i=1}^n{\sigma_i^2}}{n}\frac{\sum_{i=1}^n{z_i^2}}{\sum_{i=1}^n{\sigma_i^2 z_i^2}}$ is constant, which means $f(t)$ behaves linearly when $t$ is large enough.
  
  Exploiting a trick used in proving the Chebyshev's Sum Inequality~\citep{hardy_inequalities_1952}, we obtain $\sum_{1 \le i,j \le n}(z_i^2-z_j^2)(\sigma_i^2-\sigma_j^2) = 2n \sum_{i=1}^n{\sigma_i^2 z_i^2} - 2 (\sum_{i=1}^n{\sigma_i^2}) (\sum_{i=1}^n{z_i^2})$. The condition $\sum_{1 \le i,j \le n}(z_i^2-z_j^2)(\sigma_i^2-\sigma_j^2)>0$ thus translates into $\frac{\sum_{i=1}^n{\sigma_i^2}}{n}\frac{\sum_{i=1}^n{z_i^2}}{\sum_{i=1}^n{\sigma_i^2 z_i^2}} < 1$, which means $f(t)$ increases linearly with a slope smaller than $1$ (\emph{i.e.}, $\lim_{t\to \infty}\frac{f(t)}{t}=\frac{\sum_{i=1}^n{\sigma_i^2}}{n}\frac{\sum_{i=1}^n{z_i^2}}{\sum_{i=1}^n{\sigma_i^2 z_i^2}} < 1$).

  In summary, when $t$ approaches $0$, it is assured that $\lim_{t\to 0}f(t) > t = 0$; when $t$ is large enough, it is assured that $f(t) < t$. \textbf{Putting these two conditions together, we conclude the existence of a fixed point $t_0 > 0$ such that $f(t_0) = t_0$}.
\end{proof}

\newpage

\section{Proof of Corollary~\ref{thm:duplicate}}
\label{proof:duplicate}

Corollary~\ref{thm:duplicate}: \textup{The LogME value will remain the same if the feature consists of arbitrary replicas of the original feature. Formally speaking, if the LogME value for }$F \in \mathbb{R}^{n \times D}$ \textup{and} $y \in \mathbb{R}^{n}$ \textup{is} $\mathcal{L},$ \textup{then the LogME value for} $\tilde{F}=[F,..., F]\in R^{n\times qD}$ \textup{and} $y \in \mathbb{R}^{n}$ \textup{is also} $\mathcal{L}. \; (q \in \mathbb{N}$ \textup{is a natural number to represent the number of replicas.)} 

\begin{proof}
  Since LogME is calculated via an iterative algorithm, we prove the corollary by an iterative invariant (a quantitative relation that holds after every while-loop iteration).

  \textbf{Preliminary: SVD of $\tilde{F}$.} We have already known the SVD of $F$ is $F = U \Sigma V^T$, and $\sigma_i$ is the $i$-th largest eigenvalue of $FF^T$. Since $\tilde{F}\tilde{F}^T=qFF^T$, duplicated feature $\tilde{F} \text { has singular values } \tilde{\sigma}_{i}^{2}= \begin{cases}q \sigma_{i}^{2} & 1 \leq i \leq D \\ 0 & D+1 \leq i \leq qD \end{cases}$, and its left orthogonal matrix is the same as $F$: $\tilde{U} = U$. The right orthogonal matrix of $\tilde{F}$ is somewhat complicated. Let's find an orthogonal matrix $Q_{q\times q}$, whose entries in the first column are $\frac{1}{\sqrt{q}}$. Entries in the other columns do not matter, as long as $Q_{q\times q}$ is a valid orthogonal matrix. For example, we can use $Q_{2 \times 2} = \left[\begin{array}{cc}
    \frac{1}{\sqrt{2}} & -\frac{1}{\sqrt{2}} \\
    \frac{1}{\sqrt{2}} & \frac{1}{\sqrt{2}}  \\
    \end{array}\right]$, and $Q_{3 \times 3} = \left[\begin{array}{ccc}
      \frac{1}{\sqrt{3}} & -\frac{1}{\sqrt{6}} & -\frac{1}{\sqrt{2}} \\
      \frac{1}{\sqrt{3}} & \frac{2}{\sqrt{6}} & 0 \\
      \frac{1}{\sqrt{3}} & -\frac{1}{\sqrt{6}} & \frac{1}{\sqrt{2}}
      \end{array}\right]$. Then the right orthogonal matrix of $\tilde{F}$ is $\tilde{V}=Q_{q\times q} \otimes V \text {, where } \otimes \text { is the Kronecker product of two matrices}$. Using the block matrix form of Kronecker product, we can write down $\tilde{V}$ as $\tilde{V}=\left[\begin{array}{ccc}
        \frac{1}{\sqrt{p}} V & \ldots & \ldots \\
        \vdots & \ddots & \vdots \\
        \frac{1}{\sqrt{p}} V & \ldots & \ldots
        \end{array}\right] \in R^{q D \times q D}$, with the first $D$ columns of $\tilde{V}$ corresponding to singular values $\sqrt{q}\sigma_i, 1\le i \le D$, $\text {and the other }(q-1) \times D \text { columns of } \tilde{V}
        $ are orthogonal basis with respect to singular values $\sigma_i=0$. In summary, if the SVD of $F$ is $F = U \Sigma V^T$, then the SVD of $\tilde{F}=[F,..., F]$ is $\tilde{F} = \tilde{U} \tilde{\Sigma} \tilde{V}^T$, where $\tilde{U}=U, \tilde{\Sigma}=[\sqrt{q}\Sigma, 0, ..., 0], \tilde{V}=\left[\begin{array}{ccc}
          \frac{1}{\sqrt{q}} V & \ldots & \ldots \\
          \vdots & \ddots & \vdots \\
          \frac{1}{\sqrt{q}} V & \ldots & \ldots
          \end{array}\right] = Q_{q\times q} \otimes V$.

  \textbf{Iterative invariant:} if we apply Algorithm~\ref{alg:mackay_algorithm} to both $\tilde{F}$ and $F$, with a small change that we initialize $\tilde{\alpha} = q, \tilde{\beta}=1$, then $\tilde{\alpha}=q\alpha, \tilde{\beta}=\beta$ holds before Line~$5$. Suppose $\tilde{\alpha}=q\alpha, \tilde{\beta}=\beta$ holds before a while-loop, then we obtain:
  \begin{gather*}
    \tilde{\gamma}=\sum_{i=1}^{q D} \frac{\tilde{\beta} \tilde{\sigma}_{i}^{2}}{\tilde{\alpha}+\tilde{\beta} \tilde{\sigma}_{i}^{2}}=\sum_{i=1}^{D} \frac{q \beta \sigma_{i}^{2}}{q \alpha+q \beta \sigma_{i}^{2}} = \sum_{i=1}^{D} \frac{\beta \sigma_{i}^{2}}{\alpha+\beta \sigma_{i}^{2}}=\gamma \\
    \tilde{\Lambda}=\operatorname{diag}\left\{\tilde{\alpha}+\tilde{\beta} \tilde{\sigma}_{i}^{2}\right\}, \tilde{\alpha}+\tilde{\beta} \tilde{\sigma}_{i}^{2}= \begin{cases}q (\alpha+ \beta \sigma_{i}^{2}) & 1 \leq i \leq D \\
    q \alpha & D+1 \leq i \leq q D\end{cases} \\
  \end{gather*}

  $$
\begin{aligned}
\tilde{m} &=\tilde{\beta} \tilde{A}^{-1} \tilde{F}^{T} y=\beta \tilde{V} \tilde{\Lambda}^{-1} \tilde{V}^{T} \tilde{V} \tilde{\Sigma}^{T} \tilde{U}^{T} y=\beta \tilde{V} \tilde{\Lambda}^{-1} \tilde{\Sigma}^{T} U^{T} y \\
&=\beta\left(\left[\begin{array}{ccc}
\frac{1}{\sqrt{q}} V & \ldots & \ldots \\
\vdots & \ddots & \vdots \\
\frac{1}{\sqrt{q}} V & \ldots & \ldots
\end{array}\right]\left[\begin{array}{ll}
\frac{1}{q} \Lambda^{-1} & \\
& \frac{1}{q \alpha} I_{(q-1) \times D}
\end{array}\right]\left[\begin{array}{c}
\sqrt{q} \Sigma^{T} \\
0 \\
\ldots \\
0
\end{array}\right]\right) U^{T} y \\
&=\left[\begin{array}{c}
\frac{1}{q} V \Lambda^{-1} U^{T} y \\
\cdots \\
\frac{1}{q} V \Lambda^{-1} U^{T} y
\end{array}\right]=\left[\begin{array}{c}
\frac{1}{q} m \\
\ldots \\
\frac{1}{q} m
\end{array}\right].
\end{aligned}
$$

$
\text{Therefore }
\tilde{m}^{T} \tilde{m}=\frac{1}{q} m^{T} m, \tilde{F} \tilde{m}=[F, \ldots, F] \left[\begin{array}{c}
\frac{1}{q} m \\
\ldots \\
\frac{1}{q} m
\end{array}\right]=F m.
$

After the while-loop iteration, $\tilde\alpha'=\frac{\tilde\gamma}{\tilde{m}^T\tilde{m}}=\frac{\gamma}{\frac{1}{q}{m}^T{m}}=q\alpha', \tilde\beta'=\frac{n-\tilde\gamma}{||\tilde{F}\tilde{m}-y||_2^2}=\frac{n-\gamma}{||{F}{m}-y||_2^2}=\beta'$, then the iterative invariant $\tilde{\alpha}=q\alpha, \tilde{\beta}=\beta$ still holds. Therefore we know that when the algorithm converges, $\tilde\alpha^*=q\alpha^*, \tilde\beta^*=\beta^*$. The corresponding maximum evidence is 
\begin{equation*}
  \begin{aligned}
  \tilde{\mathcal{L}} 
  &=\frac{n}{2} \log \tilde{\beta}^{*}+\frac{q D}{2} \log \tilde{\alpha}^{*}-\frac{n}{2}\log 2 \pi-\frac{\tilde\beta^*}{2}||\tilde F\tilde m-y||_2^2-\frac{\tilde\alpha^*}{2}\tilde m^T \tilde m-\frac{1}{2} \log \left|\tilde{A}^{*}\right| \\
  &=\frac{n}{2} \log \beta^{*}+\frac{q D}{2} \log(q\alpha^{*})-\frac{n}{2}\log 2 \pi-\frac{\beta^*}{2}|| F m-y||_2^2-\frac{\alpha^*}{2}m^T m-\frac{1}{2} \log \left|\tilde{\Lambda}^{*}\right| \\
  &=\frac{n}{2} \log \beta^{*}+\frac{q D}{2} \log \left(q \alpha^{*}\right)-\frac{n}{2}\log 2 \pi-\frac{\beta^*}{2}|| F m-y||_2^2-\frac{\alpha^*}{2}m^T m \\ & \ \ \ -\frac{1}{2}\log\left|\Lambda^{*}\right|-\frac{1}{2} \log \left(q^{D} \left(q \alpha^{*}\right)^{(q-1)D}\right) \\
  &=\mathcal{L}-\frac{D}{2} \log \alpha^{*}+\frac{qD}{2}\log (q \alpha^{*})-\frac{1}{2} \log \left(q^{D}\left(q \alpha^{*}\right)^{(q-1)D}\right) \\
  &=\mathcal{L}.
  \end{aligned}
\end{equation*}

By the convergence analysis in Section~\ref{sec:convergence_analysis}, initialization of $\alpha, \beta$ only changes the initial value of $t$, which does not impact the convergence value of the fixed point iteration. Therefore, we can conclude that duplicating features will not change the value of LogME.

Although the above proof targets Algorithm~\ref{alg:mackay_algorithm}, it is straightforward to adapt the proof to Algorithm~\ref{alg:optimized_fixed_point}.
\end{proof}

\newpage

\section{Proof of Corollary~\ref{thm:padding}}
\label{proof:padding}

Corollary~\ref{thm:padding}: The \textup{LogME value will remain the same if the feature is padded with arbitrary number of zeros. Formally speaking, if the LogME value for }$F \in \mathbb{R}^{n \times D}$ \textup{and} $y \in \mathbb{R}^{n}$ \textup{is} $\mathcal{L},$ \textup{then the LogME value for} $\tilde{F}=[F, \mathbf{0}]\in R^{n\times (D+d)}$ \textup{and} $y \in \mathbb{R}^{n}$ \textup{is also} $\mathcal{L}. \; d \in \mathbb{N}$ \textup{is a natural number and } $\mathbf{0}\in \mathbb{R}^{n\times d}$  \textup{is a matrix with all zero entries.}

\begin{proof}
  The proof follows the same idea as Corollary~\ref{thm:duplicate}, but the SVD of $\tilde{F}$ is simpler than Corollary~\ref{thm:duplicate}. If the SVD of $F$ is $F = U \Sigma V^T$, then the SVD of $\tilde{F}=[F, \mathbf{0}]$ is $\tilde{F} = \tilde{U} \tilde{\Sigma} \tilde{V}^T$, where $\tilde{U} = U, \tilde{\Sigma} = [\Sigma, \mathbf{0}], \tilde{V} = \left[\begin{array}{ll}
    V & \\
    & W
    \end{array}\right], $ with $W \in \mathbb{R}^{d \times d}$ an orthogonal matrix that satisfies $W^{T} W=I_{d}$. Note that $\tilde{\Sigma} = [\Sigma, \mathbf{0}]$ translates into $\tilde{\sigma}_{i}^{2}= \begin{cases}\sigma_{i}^{2} & 1 \leq i \leq D \\ 0 & D+1 \leq i \leq D+d\end{cases}$.

    \textbf{Iterative invariant:} if we apply Algorithm~\ref{alg:mackay_algorithm} to both $\tilde{F}$ and $F$, with the same initialization $\tilde{\alpha} = 1, \tilde{\beta} = 1$, then $\tilde{\alpha}=\alpha, \tilde{\beta}=\beta$ holds before Line~$5$. Suppose $\tilde{\alpha}=\alpha, \tilde{\beta}=\beta$ holds before a while-loop, then we have:
    \begin{gather*}
      \tilde{\gamma}=\sum_{i=1}^{D + d} \frac{\tilde{\beta} \tilde{\sigma}_{i}^{2}}{\tilde{\alpha}+\tilde{\beta} \tilde{\sigma}_{i}^{2}}=\sum_{i=1}^{D} \frac{\beta \sigma_{i}^{2}}{\alpha+\beta \sigma_{i}^{2}}=\gamma \\
      \tilde{\Lambda}=\operatorname{diag}\left\{\tilde{\alpha}+\tilde{\beta} \tilde{\sigma}_{i}^{2}\right\}, \tilde{\alpha}+\tilde{\beta} \tilde{\sigma}_{i}^{2}= \begin{cases}\alpha+\beta \sigma_{i}^{2} & 1 \leq i \leq D \\
      \alpha & D+1 \leq i \leq D+d\end{cases} \\
    \end{gather*}
    \begin{equation*}
      \tilde{m}=\tilde{\beta} \tilde{A}^{-1} \tilde{F}^{T} y=\beta\left[\begin{array}{ll}
      V & \\
      & W
      \end{array}\right]\left[\begin{array}{ll}
      \Lambda^{-1} & \\
      & \frac{1}{\alpha} I_d
      \end{array}\right]\left[\begin{array}{ll}
      V^{T} & \\
      & W^{T}
      \end{array}\right]\left[\begin{array}{l}
      F^{T} \\
      \mathbf{0}_{n\times d}^{T}
      \end{array}\right] y=\left[\begin{array}{c}
      m \\
      \mathbf{0}_{d\times 1}
      \end{array}\right]
    \end{equation*}
    \begin{equation*}
      \tilde{m}^{T} \tilde{m}=m^{T} m, \tilde{F} \tilde{m}=[F, \mathbf{0}_{n\times d}]\left[\begin{array}{c}
      m \\
      \mathbf{0}_{d\times 1}
      \end{array}\right]=F m.
    \end{equation*}

    After the while-loop iteration, $\tilde\alpha'=\frac{\tilde\gamma}{\tilde{m}^T\tilde{m}}=\frac{\gamma}{{m}^T{m}}=\alpha', \tilde\beta'=\frac{n-\tilde\gamma}{||\tilde{F}\tilde{m}-y||_2^2}=\frac{n-\gamma}{||{F}{m}-y||_2^2}=\beta'$, then the iterative invariant $\tilde{\alpha}=\alpha, \tilde{\beta}=\beta$ still holds. Therefore, we know that when the algorithm converges, $\tilde\alpha^*=\alpha^*, \tilde\beta^*=\beta^*$. The corresponding maximum evidence is 
    \begin{equation*}
      \begin{aligned}
      \tilde{\mathcal{L}} 
      &=\frac{n}{2} \log \tilde{\beta}^{*}+\frac{D + d}{2} \log \tilde{\alpha}^{*}-\frac{n}{2}\log 2 \pi-\frac{\tilde\beta^*}{2}||\tilde F\tilde m-y||_2^2-\frac{\tilde\alpha^*}{2}\tilde m^T \tilde m-\frac{1}{2} \log \left|\tilde{A}^{*}\right| \\
      &=\frac{n}{2} \log \beta^{*}+\frac{D + d}{2} \log \alpha^{*}-\frac{n}{2}\log 2 \pi-\frac{\beta^*}{2}||F m-y||_2^2-\frac{\alpha^*}{2}m^T m-\frac{1}{2} \log \left|\tilde{\Lambda}^{*}\right| \\
      &=\frac{n}{2} \log \beta^{*}+\frac{D + d}{2} \log \alpha^{*}-\frac{n}{2}\log 2 \pi-\frac{\beta^*}{2}||F m-y||_2^2 -\frac{\alpha^*}{2}m^T m \\ & \ \ \ -\frac{1}{2}\log\left|\Lambda^{*}\right| -\frac{1}{2} \log \left(\alpha^{*}\right)^{d}\\
      &=\mathcal{L}+\frac{d}{2} \log \alpha^{*}-\frac{d}{2} \log \alpha^{*}  \\ &= \mathcal{L}. \\
      \end{aligned}
    \end{equation*}
\end{proof}

\newpage

\section{Detailed Descriptions of the Datasets}
\label{app:dataset}

\textbf{Aircraft:} The dataset contains fine-grained classification of 10,000 aircraft pictures which belong to 100 classes, with 100 images per class. 

\textbf{Birdsnap:} The dataset contains 49,829 images of 500 species of North American birds. 

\textbf{Caltech:} The dataset contains 9,144 pictures of objects belonging to 101 categories. There are about 40 to 800 images per category. Most categories have about 50 images.

\textbf{Cars:} The dataset contains 16,185 images of 196 classes of cars. The data is split into 8,144 training images and 8,041 testing images.

\textbf{CIFAR~10:} The dataset consists of 60,000 32$\times$32 colorful images in 10 classes, with 6,000 images per class. There are 50,000 training images and 10,000 test images.

\textbf{CIFAR~100:} The dataset is just like the CIFAR~10, except it has 100 classes containing 600 images each.

\textbf{DTD:} The dataset contains a collection of 5,640 textural images  in the wild, annotated with a series of human-centric attributes. It has 47 classes and 120 images per class.

\textbf{Pets:} The dataset contains 7,049 images of cat and dog species which belong to 47 classes, with around 200 images per class.

\textbf{SUN:} The dataset contains 39,700 scenery pictures with 397 classes and 100 samples per class.

\section{Original Results in Figures}

Original results in figures are shown in the Table~\ref{tab:figure4}, Table~\ref{tab:figure5}, and Table~\ref{tab:figure6}.

\begin{table*}[htbp]
\centering
\caption{Original results in Figure 4.}
\vskip 0.05in
\addtolength{\tabcolsep}{-4pt}
\resizebox{\textwidth}{!}{
  \begin{tabular}{clrrrrrrrrrrrrr}
    \toprule
  \multicolumn{2}{c}{task} & ResNet-34 & ResNet-50 & ResNet-101 & ResNet-152 & WideResNet-50 & DenseNet-121 & DenseNet-169 & DenseNet-201 & Inception v1 & Inception v3 & MobileNet v2 & NASNet-A Mobile & $\tau_w$ \\
  \midrule
  \multirow{5}[0]{*}{Aircraft} & Accuracy & 79.9 &  86.6 & 85.6 & 85.3 & 83.2 & 85.4 & 84.5 & 84.6 &82.7 & 88.8 &  82.8 & 72.8 & -  \\
  \cline{3-15}
  & Laplace & -2.864 & -3.127 & -3.080 & -3.158 & -3.721 & -2.235 & -1.906 & -1.754 & -2.382 & -2.822 & -2.217 & -1.481 & \textbf{-0.32} \\
  \cline{3-15}
        & LEEP  & -0.497 &  -0.412 & -0.349 & -0.308 & -0.337 & -0.431 & -0.340 & -0.462 & -0.795 & -0.492 & -0.515 & -0.506 & 0.13 \\
        \cline{3-15}
        & NCE   & -0.364 &  -0.297 & -0.244 & -0.214 & -0.248 & -0.296 & -0.259 & -0.322 & -0.348 & -0.250 & -0.411 & -0.444 & 0.39\\
        \cline{3-15}
        & LogME & 0.930 & 0.946 & 0.948 & 0.950 & 0.934 & 0.938 & 0.943 & 0.942 & 0.934 & 0.953 & 0.941 & 0.948 & \textbf{0.59} \\
        \midrule
  \multirow{4}[0]{*}{Birdsnap} & Accuracy & 59.5 &  74.7 & 73.8 & 74.3 & 63.1 & 73.2 & 71.4 & 72.6 & 73.0 & 77.2 & 69.3 & 68.3 & - \\
  \cline{3-15}
        & LEEP  & -1.758 & -1.647  & -1.553  & -1.481 & -1.554 & -1.729  & -1.756  & -1.645 & -2.483  & -1.776 & -1.951  & -1.835 &0.19 \\
        \cline{3-15}
        & NCE   & -1.640 &  -1.538  & -1.479  & -1.417  & -1.399 & -1.566  & -1.644  & -1.493  & -1.807  & -1.354 & -1.815  & -1.778  & 0.51 \\
        \cline{3-15}
        & LogME &  0.802 & 0.829  & 0.836  & 0.839 & 0.825 & 0.810  & 0.815  & 0.822  &0.806  & 0.848 & 0.808  & 0.824 & \textbf{0.66}\\
        \midrule
  \multirow{4}[0]{*}{Caltech} & Accuracy & 90.2 & 91.8 & 93.1 & 93.2 & 91.0 & 91.9 & 92.5 & 93.4 & 91.7 & 94.3 & 89.1 & 91.5 & -\\
  \cline{3-15}
        & LEEP  & -2.249 & -2.195  & -2.067  & -1.984 & -2.179 & -2.159  & -2.039  & -2.122  & -2.718  & -2.286 & -2.373  & -2.263 & 0.30 \\
        \cline{3-15}
        & NCE   & -1.899 & -1.820  & -1.777  & -1.721 & -1.828 & -1.807  & -1.774  & -1.808 & -1.849  & -1.722 & -2.009  & -1.966 & \textbf{0.69} \\
        \cline{3-15}
        & LogME &  1.362 & 1.509  & 1.548  & 1.567 & 1.505 & 1.365  & 1.417  & 1.428  & 1.440  & 1.605 & 1.365  & 1.389 & 0.66 \\
        \midrule
  \multirow{4}[0]{*}{Cars} & Accuracy &  86.4 & 91.7  & 91.7  & 92.0 & 89.7 & 91.5  & 91.5  & 91.0 & 91.0  & 92.3 & 91.0  & 88.5 & - \\
  \cline{3-15}
        & LEEP  & -1.534 & -1.570  & -1.370  & -1.334 &  -1.406 & -1.562  & -1.505  & -1.687  & -2.149  & -1.637  & -1.695  & -1.588 & 0.26\\
        \cline{3-15}
        & NCE   & -1.203 & -1.181  & -1.142  & -1.128 & -1.183 & -1.111  & -1.192  & -1.319  & -1.201  & -1.195 & -1.312  & -1.334 & 0.36 \\
        \cline{3-15}
        & LogME &  1.245 & 1.253  & 1.255  & 1.260 & 1.250  & 1.249  & 1.252  & 1.251 & 1.246  & 1.259 & 1.250  & 1.254  & \textbf{0.69}\\
        \midrule
  \multirow{4}[0]{*}{CIFAR10} & Accuracy & 97.1 &  96.8  & 97.7  & 97.9 & 97.7 & 97.2  & 97.4  & 97.4  &96.2  & 97.5  & 95.7  & 96.8 & -\\
  \cline{3-15}
        & LEEP  & -3.418 & -3.407  & -3.184  & -3.020 & -3.335 & -3.651  & -3.345  & -3.458 & -4.074  & -3.976 & -3.624  & -3.467 & 0.72\\
        \cline{3-15}
        & NCE   & -3.398 & -3.395  & -3.232  & -3.084 & -3.348 & -3.541  & -3.427  & -3.467  &-3.338  & -3.625 & -3.511  & -3.436 & 0.51\\
        \cline{3-15}
        & LogME &  0.323 & 0.388  & 0.463  & 0.469  & 0.398 & 0.302  & 0.343  & 0.369  &0.293  & 0.349 & 0.291  & 0.304  & \textbf{0.82} \\
        \midrule
  \multirow{4}[0]{*}{CIFAR100} & Accuracy & 84.5 & 84.5  & 87.0  & 87.6 & 86.4, & 84.8  & 85.0  & 86.0 & 83.2  & 86.6 & 80.8  & 83.9  & -\\
  \cline{3-15}
        & LEEP  & -3.531 & -3.520  & -3.330  & -3.167 & -3.391 & -3.715  & -3.525  & -3.643  &-4.279  & -4.100 & -3.733  & -3.560 & 0.66\\
        \cline{3-15}
        & NCE   &  -3.230 & -3.241  & -3.112  & -2.980 & -3.158 & -3.304  & -3.313  & -3.323  &-3.253  & -3.447 & -3.336  & -3.254 & 0.53 \\
        \cline{3-15}
        & LogME & 1.036 & 1.099  & 1.130  & 1.133 & 1.102 & 1.029  & 1.051  & 1.061  &1.037  & 1.070 & 1.039  & 1.051 & \textbf{0.77}  \\
        \midrule
  \multirow{4}[0]{*}{DTD} & Accuracy & 70.0 &  75.2 & 76.2 & 75.4 & 70.1 & 74.9 & 74.8 & 74.5 &73.6 & 77.2 & 72.9 & 72.8 & - \\
  \cline{3-15}
        & LEEP  &  -3.670 & -3.663  & -3.718  & -3.653 & -3.764 & -3.847  & -3.646  & -3.757  & -4.124  & -4.096 & -3.805  &  -3.691  & -0.06\\
        \cline{3-15}
        & NCE   & -3.104 & -3.119  & -3.199  &-3.138 & -3.259 & -3.198  & -3.218  & -3.203  &-3.082  & -3.261 & -3.176  & -3.149 & -0.35\\
        \cline{3-15}
        & LogME & 0.704 & 0.761  & 0.757  & 0.766 & 0.731 & 0.710  & 0.730  & 0.730  &0.727  & 0.746 & 0.712  & 0.724  & \textbf{0.50}\\
        \midrule
  \multirow{4}[0]{*}{Pets} & Accuracy & 92.3 & 92.5  & 94.0  & 94.5 & 92.8 & 92.9  & 93.1  & 92.8  & 91.9  & 93.5 & 90.5  &  89.4 & -\\
  \cline{3-15}
        & LEEP  & -1.174 & -1.031  & -0.915  & -0.892 & -0.945 & -1.100  & -1.111  & -1.108  &-1.520  & -1.129  &-1.228  & -1.150 & 0.66\\
        \cline{3-15}
        & NCE   &  -1.094 & -0.956  & -0.885  & -0.862 & -0.900 & -0.987  & -1.072  & -1.026  & -1.076  & -0.893 & -1.156  & -1.146 & \textbf{0.83}\\
        \cline{3-15}
        & LogME &  0.835 & 1.029  & 1.061  & 1.084 & 1.016 & 0.839  & 0.874  & 0.908  &0.913  & 1.191  & 0.821  & 0.833 & 0.61\\
        \midrule
  \multirow{4}[0]{*}{SUN} & Accuracy & 63.1 & 64.7  & 64.8  & 66.0 & 67.4 & 62.3  & 63.0  & 64.7  & 62.0  & 65.7 & 60.5  & 60.7 & -\\
  \cline{3-15}
        & LEEP  & -2.727 & -2.611  & -2.531  & -2.513 &  -2.569 & -2.713  & -2.570  & -2.618 & -3.153  & -2.943 & -2.764  & -2.687 & 0.54\\
        \cline{3-15}
        & NCE   &  -2.573 & -2.469  & -2.455  & -2.444 & -2.457 & -2.500  & -2.480  & -2.465  & -2.534  & -2.529 & -2.590  & -2.586  & 0.68\\
        \cline{3-15}
        & LogME & 1.704 & 1.744  & 1.749  & 1.755 & 1.750 & 1.704  & 1.716  & 1.718  & 1.715  & 1.753 & 1.713  & 1.721 & \textbf{0.71}\\
        \bottomrule
  \end{tabular}%
}
\label{tab:figure4}%
\end{table*}%

\begin{table*}[htbp]
\centering
\caption{Original results in Figure 5.}
\vskip 0.05in
\addtolength{\tabcolsep}{-4pt}
\resizebox{\textwidth}{!}{
  \begin{tabular}{clrrrrrrrrrrrrr}
    \toprule
  \multicolumn{2}{c}{task} & ResNet-34 & ResNet-50 & ResNet-101 & ResNet-152 & WideResNet-50 & DenseNet-121 & DenseNet-169 & DenseNet-201 & Inception v1 & Inception v3 & MobileNet v2 & NASNet-A Mobile & $\tau_w$ \\
  \midrule
  \multirow{2}[0]{*}{dSprites} & MSE & 0.037 &  0.031   &  0.028  &   0.028  & 0.034  &  0.039     &   0.035    &   0.036    &   0.045    &  0.044     &   0.037    &  0.035     &  - \\
  \cline{3-15}
        & LogME &  1.05 & 1.53  &    1.64   &   1.63  &  1.31 &   1.35    &    1.25   &   1.34    &  1.18     &  1.22     &   1.18    &   1.39    & 0.79 \\
        \bottomrule
  \end{tabular}%
}
\label{tab:figure5}%
\end{table*}%

\begin{table*}[htbp]
\centering
\caption{\small Original results in Figure~\ref{fig:nlp}. (Popularity is measured by download count in millions.)}
\vskip 0.05in
\resizebox{\textwidth}{!}{
  \begin{tabular}{clrrrrrrrrr}
    \toprule
  \multicolumn{2}{c}{task} & RoBERTa & RoBERTa-D & uncased BERT-D & cased BERT-D & ALBERT-v1 & ALBERT-v2 & ELECTRA-base & ELECTRA-small & $\tau_w$ \\
  \midrule
  \multirow{3}[0]{*}{MNLI} & Accuracy &  87.6    & 84.0  & 82.2 & 81.5 & 81.6& 84.6& 79.7& 85.8& -     \\
  \cline{3-11}
        & LogME &  -0.568   &   -0.599    &   -0.603    &  -0.612&  -0.614&-0.594&  -0.666&-0.621 & 0.66 \\
  \cline{3-11}
        & Popularity &  3.78   &   0.61    &   6.01    &  1.09 &  0.11 & 1.25 &  0.13 & 0.23 & 0.28 \\
        \midrule
  \multirow{3}[0]{*}{QQP} & Accuracy & 91.9 & 89.4 & 88.5 & 87.8    &  -     &  -     &   -    &    - & -   \\
  \cline{3-11}
        & LogME &  -0.465 &   -0.492  &  -0.488  &  -0.521            &   -    & -      &    -   &      - &  0.73\\
  \cline{3-11}
        & Popularity &  3.78   &   0.61    &   6.01    &  1.09 &  - & - &  - & - & 0.00 \\
        \midrule
  \multirow{3}[0]{*}{QNLI} & Accuracy &  92.8 & 90.8 & 89.2 & 88.2   &  -     &    -   &     -  &    -   &    -   \\
  \cline{3-11}
        & LogME &  -0.565    &  -0.603    &   -0.613    &   -0.618            & -      &      - &   -    &     -  & 1.00 \\
  \cline{3-11}
        & Popularity &  3.78   &   0.61    &   6.01    &  1.09 &  - & - &  - & - & 0.00 \\
        \midrule
  \multirow{3}[0]{*}{SST-2} & Accuracy &    94.8 & 92.5 & 91.3 & 90.4 & 90.3 & 92.9   & - & - & -    \\
  \cline{3-11}
        & LogME &   -0.312 & -0.330   &   -0.331    &    -0.353& -0.525&-0.447       &    -   &   -    & 0.68 \\
  \cline{3-11}
        & Popularity &  3.78   &   0.61    &   6.01    &  1.09 &  0.11 & 1.25 &  - & - & 0.38 \\
        \midrule
  \multirow{3}[0]{*}{CoLA} & Accuracy &  63.6 & 59.3 & 51.3 & 47.2  & - &- &- &- & -  \\
  \cline{3-11}
        & LogME &   -0.499   &    -0.536   &   -0.568    &  -0.572            &  -     &   -    &   -    &     -  & 1.00 \\
  \cline{3-11}
        & Popularity &  3.78   &   0.61    &   6.01    &  1.09 &  - & - &  - & - & 0.00 \\
        \midrule
  \multirow{3}[0]{*}{MRPC} & Accuracy &   90.2& 86.6 & 87.5 & 85.6 &- &- &- &- &-    \\
  \cline{3-11}
        & LogME &  -0.573  &   -0.586    & -0.605      &    -0.604           &   -    &     -  &   -    &    -   &  0.53\\
  \cline{3-11}
        & Popularity &  3.78   &   0.61    &   6.01    &  1.09 &  - & - &  - & - & 0.33 \\
        \midrule
  \multirow{3}[0]{*}{RTE} & Accuracy & 78.7 & 67.9& 59.9    & 60.6      &    -   &  -     &  -     & -      &   -    \\
  \cline{3-11}
        & LogME &    -0.709 &  -0.723     &    -0.725   &    -0.725           &   -    &  -     &   -    &    -   &  1.00\\
  \cline{3-11}
        & Popularity &  3.78   &   0.61    &   6.01    &  1.09 &  - & - &  - & - & -0.29 \\
  \bottomrule
  \end{tabular}%
}
\label{tab:figure6}%
\end{table*}%

\section{Complete Results in Prompt Learning}

\begin{table}[htbp]
  \centering
  \caption{Complete results in prompt learning with manually selected anchor words.}
  \vskip 0.05in
    \begin{tabular}{l|r|r|r|r|r}
      \toprule
    \diagbox{anchor P}{accuracy}{anchor N}  & negative & bad & ill & evil & poor \\
    \hline
    positive & 49.8  & 52.8  & 49.1  & 49.1  & 60.9 \\
    good  & 51.0    & 50.9  & 49.0    & 52.4  & 50.9 \\
    fine  & 51.7  & 51.0    & 49.1  & 54.5  & 50.9 \\
    great & 55.4  & 53.1  & 49.1  & \textbf{61.4}  & 51.0 \\
    nice  & 51.6  & 50.6  & 49.1  & 51.0    & 50.8 \\
    \bottomrule
    \end{tabular}%
  \label{tab:prompt_learning}%
\end{table}%

\newpage

\section{Full Figure in Convergence Analysis}

\begin{figure}[htbp]
  \centering
  \includegraphics[width=\columnwidth]{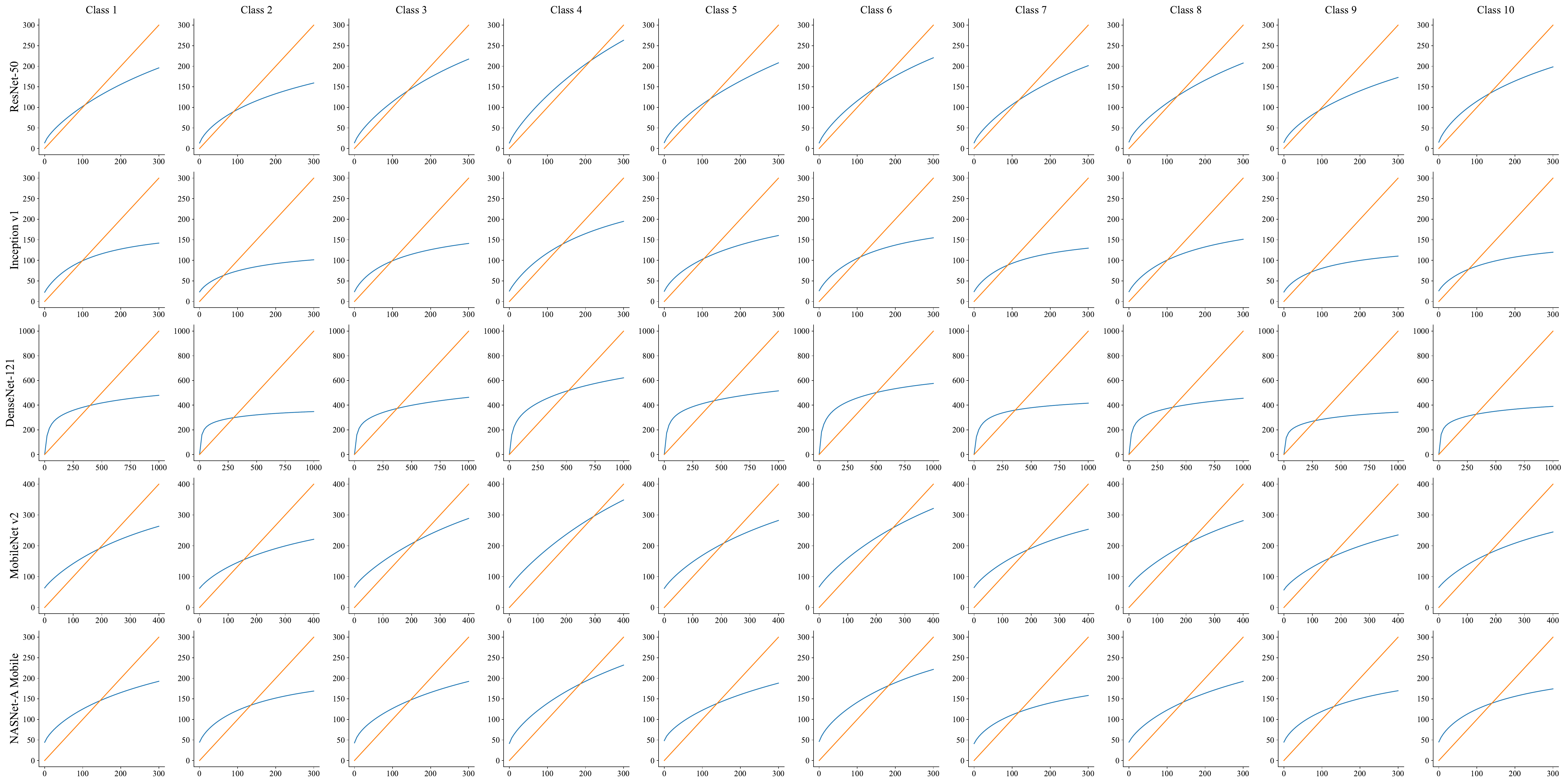}
  \includegraphics[width=.3\columnwidth]{conv_legend.pdf}
  \caption{Fixed points of $f(t)$ in Equation~\ref{eq:fixpoint} for all $10$ classes in CIFAR10 with $5$ pre-trained models. We plot $t' = f(t)$ (in blue) and $  t' = t$ (in orange), whose intersections are fixed points. The existence of fixed points guarantees the convergence of the evidence maximization procedure in LogME.}
  \label{fig:full}
\end{figure}

\end{document}